\documentclass[review,authoryear]{article}
\pdfoutput=1
\usepackage{natbib}
\usepackage{graphicx} 
\usepackage{amsmath}
\usepackage{amssymb}
\usepackage{array}
\usepackage{color}
\usepackage{rotating}
\usepackage{multirow}
\usepackage[english,ruled,vlined]{algorithm2e}
\usepackage{rotating}
\usepackage{wrapfig}
\usepackage[text={6.6in,8.7in},centering]{geometry}
\usepackage{url}
\usepackage{comment}
\usepackage{makecell}

\newtheorem{defin}{Definition}
\newcommand{\commentout}[1]{}
\newcommand{\shortversion}[1]{}

\newcommand{\includebigtable}[1]{}
\newcommand{\iurow}{IUR}

\newcommand{\class}[1]{{\sf #1}}
\newcommand{\slot}[2]{{{\sf #1}.{\em #2}}}
\newcommand{\attribute}[1]{{\em #1}}
\newcommand{\slotchainset}[2]{{{\sf #1}.{\bf\em #2}}}
\newcommand{\slotchain}[3]{{\slotchainset{#1}{#2}.{\em #3}}}
\newcommand{\slotprime}[2]{{{\sf #1$'$\!}.{\em #2}}}
\newcommand{\objectprime}[2]{\lowercase{{\em #1$'$\!}.{\em #2}}}
\newcommand{\objectslotprime}[2]{\lowercase{{\em #1$'$\!}}.{\bf\em #2}}
\newcommand{\slotprimechainset}[2]{{{\sf #1$'$\!}.{\bf\em #2}}}
\newcommand{\slotprimechain}[3]{{\slotchainset{#1$'$\!}{#2}.{\em #3}}}

\title{Relational Approach to Knowledge Engineering for POMDP-based Assistance Systems as a Translation of a Psychological Model}

\author{Marek Grze\'{s},  Jesse Hoey and Shehroz Khan\\
School of Computer Science, University of Waterloo, Canada\\
\tt{\{mgrzes, jhoey, s255khan\}@cs.uwaterloo.ca}
\and
Alex Mihailidis and Stephen Czarnuch\\
Department of Occupational Science and Occupational Therapy, University of Toronto, Canada\\
\and
Dan Jackson\\
School of Computing Science, Newcastle University, UK\\
\and
Andrew Monk\\
Department of Psychology, University of York, UK\\
}

\begin{document}

\maketitle

\begin{abstract}
Assistive systems for persons with cognitive disabilities (e.g. dementia) are difficult to build due to the wide range of different approaches people can take to accomplishing the same task, and the significant uncertainties that arise from both the unpredictability of client's behaviours and from noise in sensor readings.  
Partially observable Markov decision process (POMDP) models have been used successfully as the reasoning engine behind such assistive systems for  small multi-step tasks such as hand washing. POMDP models are a powerful, yet flexible framework for modelling assistance that can deal with uncertainty and utility. Unfortunately, POMDPs usually require a very labour intensive, manual procedure for their definition and construction. Our previous work has described a knowledge driven method for automatically generating POMDP activity recognition and context sensitive prompting systems for complex tasks. We call the resulting POMDP a SNAP (SyNdetic Assistance Process). The spreadsheet-like result of the analysis does not correspond to the POMDP model directly and the translation to a formal POMDP representation is required. To date, this translation had to be performed manually by a trained POMDP expert. In this paper, we formalise and automate this translation process using a probabilistic relational model (PRM) encoded in a relational database. The database encodes the relational skeleton of the PRM, and includes the goals, action preconditions, environment states, cognitive model, client and system actions (i.e., the outcome of the SNAP analysis), as well as relevant sensor models.  The database
is easy to approach for someone who is not an expert in POMDPs, allowing them to fill in the necessary details of a task using a simple and intuitive procedure. 
The database, when filled, implicitly defines a ground instance of the relational skeleton, which we extract using an automated procedure, thus generating a POMDP model of the assistance task.  A strength of the database is that it allows constraints to be specified, such that we can verify the POMDP model is, indeed, valid for the task given the analysis.   We demonstrate the method by eliciting three assistance tasks from non-experts: handwashing, and toothbrushing for elderly persons with dementia, and on a factory assembly task for persons with a cognitive disability.  We validate the resulting POMDP models using case-based simulations to show that they are reasonable for the domains. We also show a complete case study of a designer specifying one database, including an evaluation in a real-life experiment with a human actor. 
\end{abstract}


\section{Introduction}
\label{sec:intro}
Quality of life (QOL) of persons with a cognitive disability (e.g. dementia, developmental disabilities) is increased significantly if they can engage in `normal' routines in their own homes, workplaces, and communities.  However, they generally require some assistance in order to do so. For example, difficulties performing common activities such as self-care or work-related tasks, may trigger the need for personal assistance or relocation to residential care settings~\citep{Gill03}. Moreover, it is associated with diminished QOL, poor self-esteem, anxiety, and social isolation for the person and their caregiver~\citep{Burns00}. 

Technology to support people in their need to live independently is currently available in the form of personal and social alarms and environmental adaptations and aids. Looking to the future, we can imagine intelligent, pervasive computing technologies using sensors and effectors that help with more difficult cognitive problems in planning, sequencing and attention. A key problem in the construction of such intelligent technologies is the automatic analysis of people's behaviours from sensory data. Activities need to be recognized and, by incorporating domain specific expert knowledge, reasonable conclusions have to be drawn which ultimately enables the environment to perform appropriate actions through a set of actuators. In the example of assisting people with dementia, the smart environment would prompt (i.e., issue a voice or video prompt) whenever the clients get stuck in their activities of daily living.

\commentout{
Technology to support people in their need to live independently is currently available in the form of personal and social alarms and environmental adaptations and aids. Looking to the future, we can imagine intelligent, pervasive computing technologies using sensors and effectors that help with more difficult cognitive problems in planning, sequencing and attention. In the example of assisting people with dementia, the smart environment would prompt whenever the residents get stuck in their activities of daily living.
}

The technical challenge of developing useful prompts and a sensing and modelling system that allows them to be delivered only at the appropriate time is difficult, due to issues such as the system needing to be able to determine the type of prompt to provide, the need for the system to recognize changes in the abilities of the person and adapt the prompt accordingly, and the need to give different prompts for different sequences within the same task.  However, such a system has been shown to be achievable through the use of advanced planning and decision making approaches. One of the more sophisticated of these types of systems is the COACH~\citep{Hoey10b}. COACH uses computer vision to monitor the progress of a person with dementia washing their hands and prompts only when necessary. COACH uses a partially observable Markov decision process (POMDP), a temporal probabilistic model that represents a decision making process based on environmental observations.  The COACH model is flexible in that it can be applied to different tasks~\citep{Hoey11a}. However, each new task requires substantial re-engineering and re-design to produce a working assistance system, which currently requires massive expert knowledge for generalization and broader applicability to different tasks. An automatic generation of such prompting systems would substantially reduce the manual efforts necessary for creating assistance systems, which are tailored to specific situations and tasks, and environments.  In general, the use of {\em a-priori} knowledge in the design of assistance systems is a key unsolved research question.  Researchers have looked at specifying and using ontologies~\citep{Chen08}, information from the Internet~\citep{Pentney08}, logical knowledge bases~\citep{Chen08,Mastrogiovanni08}, and programming interfaces for context aware human-computer interaction~\citep{Salber99}.  

In our previous work, we have developed a knowledge driven method for automatically generating POMDP activity recognition and context sensitive prompting systems~\citep{SnapPMC11}.  The approach starts with a description of a task and the environment in which it is to be carried out that is relatively easy to generate. Interaction Unit (IU) analysis \citep{RyuMonk09}, a psychologically motivated method for transcoding interactions relevant for fulfilling a certain task, is used for obtaining a formalized, i.e., machine interpretable task description. This is then combined with a specification of the available sensors and effectors to build a working model that is capable of analysing ongoing activities and issuing prompts.  We call the resulting model a SyNdetic Assistance Process or SNAP. However, the current system uses an ad-hoc method for transcoding the IU analysis into the POMDP model.  While each of the factors are well defined, fairly detailed and manual specification is required to enable the translation.

The long-term goal of the approach presented in this paper is to allow end-users, such as health professionals, caregivers, and family members, to specify and develop their own context sensitive prompting systems for their needs as they arise. This paper describes a step in this direction by proposing a probabilistic relational model (PRM)\citep{getoor07prm} defined as a relational database that encodes a domain independent relational dynamic model and serves to mediate the translation between the IU analysis and the POMDP specification. The PRM encodes the constraints required by the POMDP in such a way that, once specified, the database can be used to generate a POMDP specification automatically that is guaranteed to be valid (according to the SNAP model).  The PRM serves as a schema that can be instantiated for a particular task using a simple and intuitive specification method.  The probabilistic dependencies in the PRM are boiled down to a small set of parameters that additionally need to be specified to produce a working POMDP-based assistance system.
This novel approach helps in coping with a number of issues, such as validation, maintenance, structure, tool support, association with a workflow method etc., which were identified to be critical for tools and methodologies that could support knowledge engineering in planning \citep{mccluskey00ke4planning_readmap}. This paper gives the details of this relational database, and then demonstrates the application of the method to specify a POMDP in three examples: two are for building systems to assist persons with dementia during activities of daily living, and one is to assist persons with Down's syndrome during a factory assembly task.  We show how the method requires little prior knowledge of POMDPs, and how it makes specification of relatively complex tasks a matter of a few hours of work for a single coder.

The remainder of this paper is structured as follows. First, we give an overview of the basic building blocks: POMDPs, the IU analysis, knowledge engineering, and probabilistic relational models (PRMs). Then, Section~\ref{SSEC:RD} describes the specific PRM and relational database that we use, and shows how the database can be leveraged in the translation of the IU analysis to a POMDP planning system. In Section~\ref{SEC:SYSUSE}, we show a case study that explains step-by-step the design process of an example prompting system. Section~\ref{sec:experiments} shows how the method can be applied to three tasks, including the real-file simulation with a human actor, and then the paper concludes.

This paper is describing assistive systems to help persons with a cognitive disability. Throughout the paper, we will refer to the person requiring assistance as the {\em client}, and to any person providing this assistance as the {\em caregiver}.  A third person involved is the {\em designer}, who will be the one using the system we describe in this paper to create the assistive technology. Thus, our primary target user in this paper is the {\em designer}.

\section{Overview of Core Concepts}\label{sec:overview}

This section reviews the core concepts necessary to fully explain our method.  
We start by introducing the concept of POMDPs (Section~\ref{SSEC:POMDP}) which is the core mathematical model and represents the final outcome of our modelling task, i.e., the actual machine-readable specification of the prompting system.  Next, we introduce a method that is used in order to create the initial SNAP specification for a new task. The outcome of this step is a spreadsheet-like document that describes the task (Section~\ref{sec:iu}).   Section~\ref{SSEC:SNAPDB} then reviews key concepts of knowledge engineering, and motivates the primary objective of this paper: to provide a link between the task analysis of Section~\ref{sec:iu} and the POMDP model.  Finally, Section~\ref{SSEC:PRM} overviews probabilistic relational models (PRMs).  We use a PRM as our primary abstraction of the POMDP model, enabling designers to specify POMDP models at a level of abstraction that is appropriate. 

\subsection{Partially observable Markov decision processes}\label{SSEC:POMDP}

\begin{figure}[htb]
\begin{center}
\includegraphics[width=0.6\textwidth]{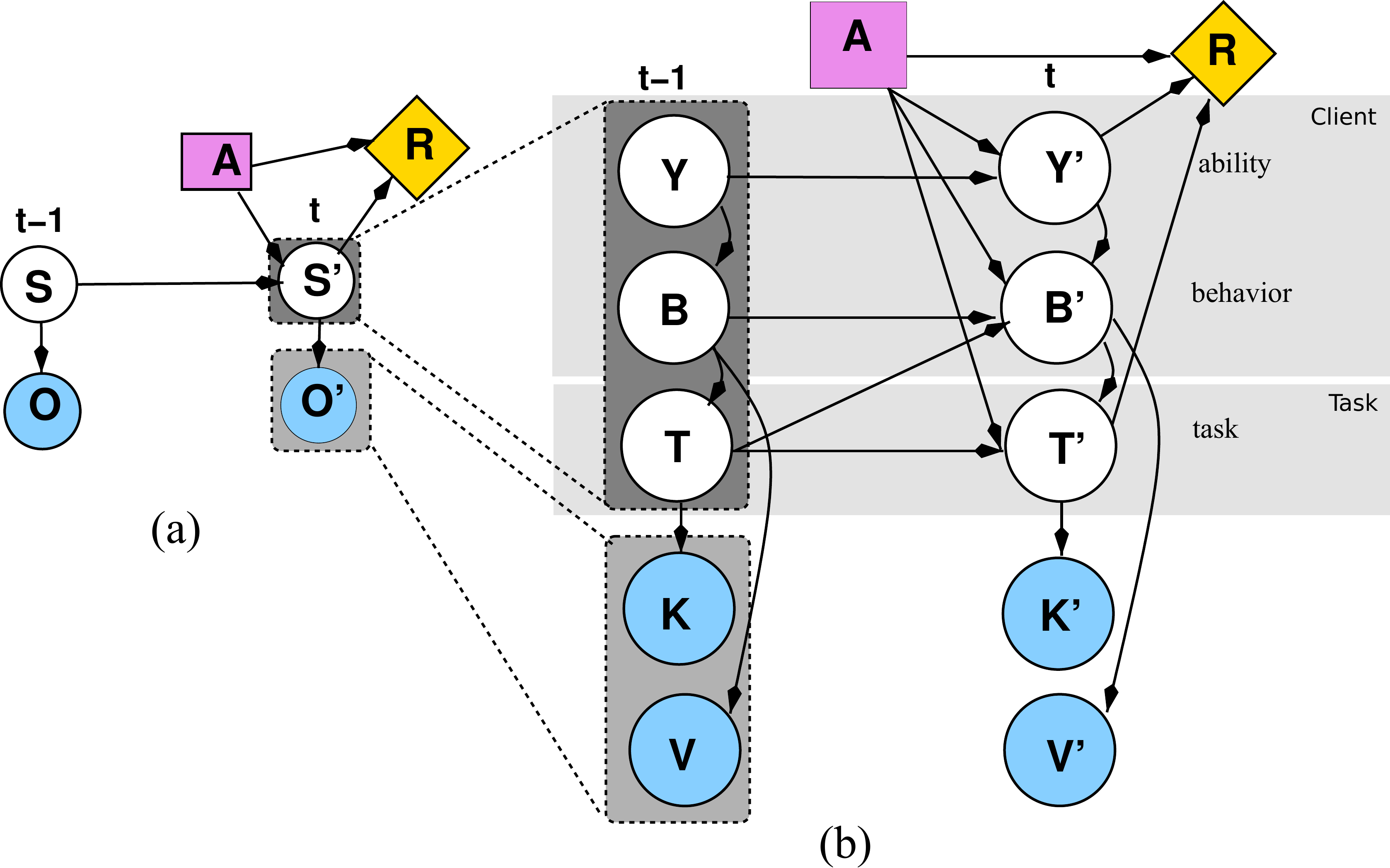} 
\end{center}
\caption{Two time slices of (a) a general POMDP; (b) a factored POMDP for interactions with assistive technology. $T$ is the set of task variables, $B$ represents behaviours of the client, and $Y$ client's abilities. $B$ and $Y$ constitute the model of the client. $A$ is the action of the system, $R$ the reward, and $K$ and $V$ are observations/sensors for task variables and client's behaviours correspondingly. Primed variables represent the next time step of the dynamic model and arrows indicate probabilistic dependence like in standard Bayesian networks.
}
\label{fig:pomdp}
\end{figure}

A POMDP is a probabilistic temporal model of a system interacting with its environment~\citep{astrom,poupart11pompds}, and is described by (1) a finite set of state variables, the cross product of which gives the state space, $S$; (2) a set of observation variables, $O$ (the outputs of some sensors); (3) a set of system actions, $A$; (4) a reward function, $R(s,a,s')$, giving the relative utility of transiting from state $s$ to $s'$ under action $a$; (5) a stochastic transition model $Pr:S\times A\rightarrow\Delta S$ (a mapping from states and actions to distributions over states), with $Pr(s'|s,a)$ denoting the probability of moving from state $s$ to $s'$ when action $a$ is taken; and (6) a stochastic observation model with $Pr(o|s)$ denoting the probability of making observation $o$ while the system is in state $s$. Figure~\ref{fig:pomdp}(a) shows a POMDP as a Dynamic Bayesian network (DBN) with actions and rewards, where arrows are interpretable as causal links between variables.  

A POMDP for personal assistance breaks the state space down into three key factors as shown in Figure~\ref{fig:pomdp}(b): states describing elements of the functional task in the real world, $T$, e.g. whether the water has been boiled or not (the `task factor model'), states capturing the client's cognitive capacities, $Y$, e.g., to remember what they are supposed to do next (the `ability factor model'), and states capturing an inferred history of what the client has actually done since the last update, $B$, e.g. fill the kettle (the `behaviour factor model'). We use the word `behaviour' here to describe actions of the client to distinguish them from actions of the system (i.e., prompts to the client). A preliminary version of this model was explored and used in different contexts in~\citep{Hoey05b,Mihailidis08,Hoey10b}.  As we will see, these three factors relate to the psychological analysis, and keeping them separate will ease the translation between the two.

The observations $O$ and $O'$ are states of the sensors at times $t-1$ and $t$. It is convenient to divide $O$ into the states of sensors relevant to the task environment ($K$) and states of sensors relevant to client behaviour ($V$). Whereas the first type of sensors monitors, for example, movements of objects in the environment, the latter corresponds to virtual sensors that, for example, monitor the activities the acting person pursues. The system's actions are various prompts or memory aids the system can use to help a client remember things in the task. For persons with dementia, simple audio prompts are often very effective, and are used in the COACH system along with video demonstrations~\citep{Mihailidis08,Hoey10b}. Finally, the observations are any information from sensors in the environment that can give evidence to the system about the client's behaviours or the task. 

A POMDP defines a mathematical representation of a given, partially observable, stochastic decision problem. Our aim in this paper is to provide an automatic translation of the IU analysis of assistance tasks into such a mathematical description. Once such a POMDP model is specified, it can be solved using an arbitrary POMDP planner in order to obtain a policy of action, i.e., to determine in which states of the environment prompts should be issued. Thus our work in this paper does not focus on solving POMDPs (i.e. we do not propose a new POMDP planning algorithm), rather on methodologies for specifying POMDPs that fall into the category of knowledge engineering for planning.

\subsection{Specifying the task: Interaction Unit Analysis}\label{sec:iu}

Task analysis has a long history in Human Factors~\citep{Kirwan92} where this approach is typically used to help define and break-down `activities of daily living' (ADL) -- i.e. activities that include self-care tasks, household duties, and personal management such as paying bills. The emphasis in task analysis is on describing the actions taken by a client and the intentions (goals and sub-goals) that give rise to those actions. There has been less emphasis on how actions are driven by the current state or changes in the environment. Syndetic modeling~\citep{Duke98} remedies this omission by describing the conjunction of cognitive and environmental precursors for each action. Modelling both cognitive and environmental mechanisms at the level of individual actions turns out to be much more efficient than building separate cognitive and environmental models~\citep{RyuMonk09}.

 
The task analysis technique~\citep{WhertonMonk10} breaks a task down into a set of goals, states, abilities and behaviours, and defines a hierarchy of tasks that can be mapped to a POMDP, a policy for which will be a situated prompting system for a particular task~\citep{SnapPMC11}. The technique involves an experimenter video-taping a person being assisted during the task, and then transcribing and analysing the video. The end-result is an Interaction Unit (IU) analysis that uncovers the states and goals of the task, the client's cognitive abilities, and the client's actions. A simplified example for the first step in tea-making (getting out the cup and putting in a tea-bag) is shown in Table~\ref{tab:iuanal}. The rows in the table show a sequence of steps, with the client's current goals, the current state of the environment, the abilities that are necessary to complete the necessary step, and the behaviour that is called for. The abilities are broken down into ability to {\em recall} what they are doing, to {\em recognise} necessary objects like the kettle, and to perceive {\em affordances} of the environment.
\begin{table}[htbp]
\begin{center}
{\small
\begin{tabular}{|l|l|l|l|l|}
\cline{1-5}
{\bf IU} & {\bf Goals} & {\bf Task States} & {\bf Abilities} & {\bf Behaviours} \\ \cline{1-5}\hline
 1 & Final & cup empty on tray, box~closed & Rn cup on tray, Rl step & No Action\\ \cline{1-5}
2 & Final, cup~TB & cup~empty on~tray, box~closed  & Af cup on tray WS & Move cup tray$\rightarrow$WS\\ \cline{1-5}
 3 & Final, cup TB & cup empty on WS, box~closed & Rl box contains TB                & Alter box to open\\
   &               &                             &                     Af~box~closed &                  \\ \cline{1-5}
 4 & Final, cup TB & cup empty on WS, box~open & Af TB in box cup & Move TB box$\rightarrow$cup\\ \cline{1-5}
 5 & Final  & cup tb on WS, box open & Af box open & Alter box to closed\\ \cline{1-5}
  & Final & cup tb on WS, box closed &  & \\ \hline
\end{tabular}
}
\caption{IU analysis of the first step in tea making. The first column provides an index of a specific interaction unit also referred to as a row in the IU table. The second column contains the goal the client is trying to achieve in a specific row, whereas the third column defines the set of task states which are relevant in the row. The fourth column lists abilities which are necessary for the behaviour shown in the fifth column to happen. Types of behaviours are distinguished by the prefix of their names, e.g., Rn=recognition, Rl=Recall, Af=Affordance. The remaining shorthand notation is: tb=teabag, ws=work surface.}
\label{tab:iuanal}
\end{center}
\end{table}

A second stage of analysis involves proposing a set of sensors and actuators that can be retrofitted to the client's environment for the particular task, and providing a specification of the sensors that consists of three elements: (1) a name for each sensor and the values it can take on (e.g. on/off); (2) a mapping from sensors to the states and behaviours in the IU analysis showing the evidentiary relationships, and (3) measurements of each sensor's reliability at detecting the states/behaviours it is related to in the mapping.

\begin{figure}
 \centering
 \includegraphics[scale=0.6]{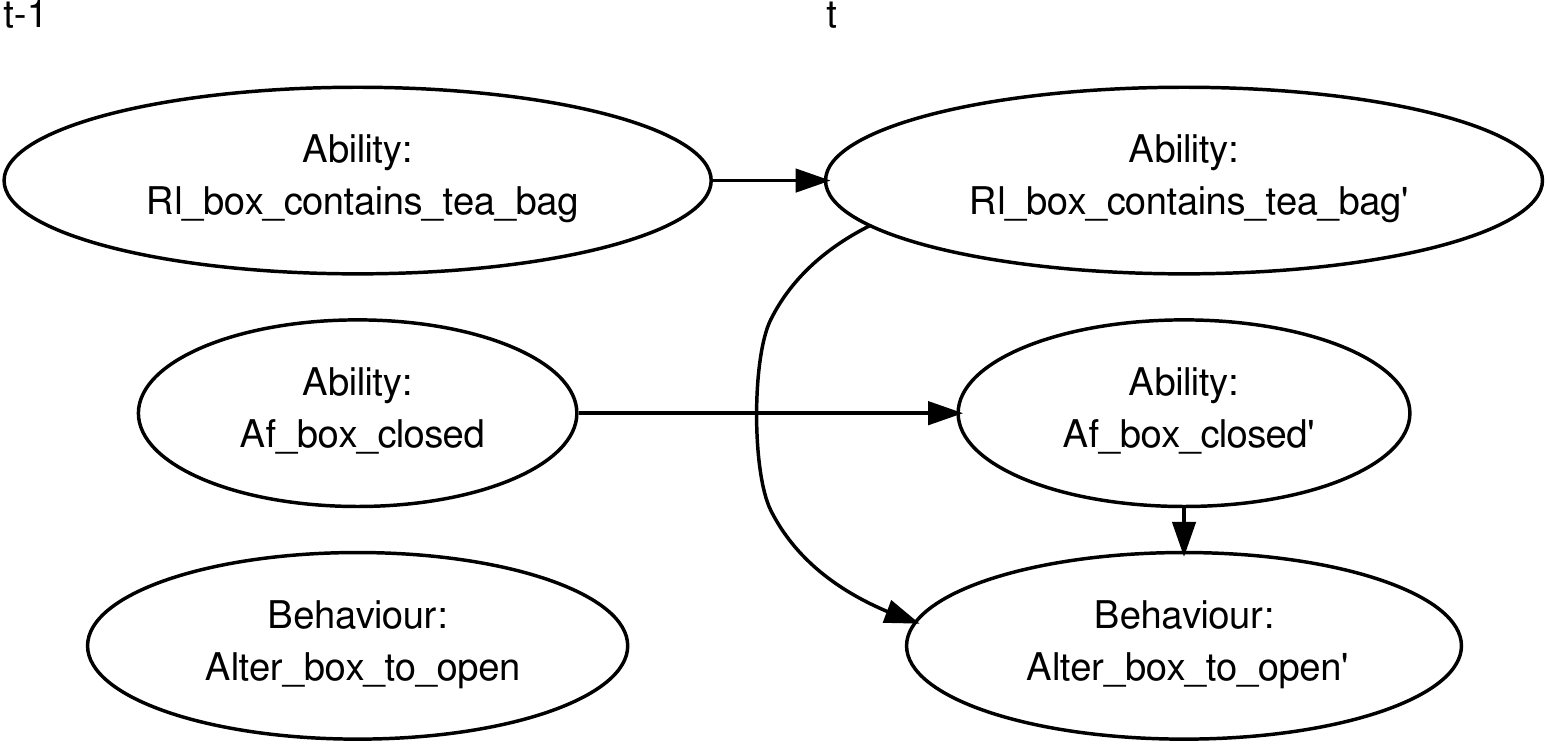}
\caption{Example probabilistic dependencies read from the IU table specified in Table~\ref{tab:iuanal}. The example corresponds to row 3 in Table~\ref{tab:iuanal} that specifies that behaviour Alter\_box\_to\_open requires abilities Rl\_box\_contains\_tea\_bag and Af\_box\_closed.}
\label{FIG:EX:TEA}
\end{figure}

The IU analysis (e.g.~Table~\ref{tab:iuanal}), along with the specification of sensors which are part of SNAP can be converted to a POMDP model by factoring the state space as shown in Figure~\ref{fig:pomdp}(b). The method is described in detail in~\citep{SnapPMC11}; here we give a brief overview. The {\em task} variables are a characterisation of the domain in terms of a set of high-level variables, and correspond to the entries in the task states column in Table~\ref{tab:iuanal}. For example, in the first step of tea making, these include the box condition (open, closed) and the cup contents (empty or with teabag). The task states are changed by the client's {\em behaviour}, $B$, a single variable with values for each behaviour in Table~\ref{tab:iuanal}. For the first IU group in tea making, these include opening/closing the box, moving the teabag to the cup, and doing nothing or something unrelated (these last two behaviours are always present). The clients' {\em abilities} are their cognitive states, and model the ability of the client to recall (Rl), recognise (Rn) and remember affordances (Af). For the first IU group, these include the ability to recognise the tea box and the ability to perceive the affordance of moving the teabag to the cup. Example DBN relationships read from Table~\ref{tab:iuanal} are shown in Figure~\ref{FIG:EX:TEA}. The example corresponds to row 3 in Table~\ref{tab:iuanal} that specifies that behaviour Alter\_box\_to\_open requires abilities Rl\_box\_contains\_tea\_bag and Af\_box\_closed.

The system actions are prompts that can be given to help the client regain a lost ability. We define one system action for each necessary ability in the task. The actions correspond to a prompt or signal that will help the client with this particular ability, if missing. System actions (prompts) are automatically derived from the IU analysis because abilities of different types are specified by the person performing the analysis, and the system can then automatically generate one system action (prompt) for every ability.

{\em Task} and {\em behaviour} variables generate observations, $O=\{K,V\}$. For example, in a kitchen environment there may be sensors in the counter-tops to detect if a cup is placed on them, sensors in the teabags to detect if they are placed in the cup, and sensors in the kettle to detect `pouring' motions. The sensor noise is measured independently (as a miss/false positive rate for each state/sensor combination)~\citep{Pham09,SnapPMC11}.

Some of the required dynamics (i.e., behaviour relevance) and initial state are produced directly from the IU analysis (Table~\ref{tab:iuanal}), by looking across each row and associating state transitions between rows. We take this to be deterministic, as any uncertainty will be introduced by the client's abilities (so we assume a perfectly able client is able to always successfully complete each step). Each prompting action improves its associated cognitive ability. For example, the `prompt recognition cup' action (e.g a light shone on the cup) makes it more likely that the client can recognise the cup if they cannot do so already.  The reward function specifies the goal states (in Table~\ref{tab:iuanal}), and assigns a cost to each prompt, as client independence is paramount.

\subsection{Requirements from Knowledge Engineering}\label{SSEC:SNAPDB}

The IU analysis and the sensor specification need to be translated into a POMDP model, and then the policy of action can be generated. The relational database provides a natural link between these two elements of the prompting system, and the use of the database represents additionally a novel approach to knowledge engineering (KE) for planning. For an extensive review of the challenges which KE for planning faces, the reader is referred to \citep{mccluskey00ke4planning_readmap}. This area is essentially investigating the problem of how planning domain models can be specified by technology designers who are not necessarily familiar with the AI planning technology. In \citep{mccluskey00ke4planning_readmap}, authors collected a number of requirements which such a methodology should satisfy. Some of most important ones are: (1) acquisition, (2) validation, (3) maintenance, and additionally the representation language should be: (4) structured, (5) associated with a workflow method, (6) easy to assess with regard to the complexity of the model, (7) tool supported, (8) expressive and customizable, and (9) with a clear syntax and semantics. In our work on the SNAP process, we found that these requirements can be, to a great extent, supported when one applies the relational database formalism to store and to process the domain model. The acquisition step (1) does not have its full coverage in our case since, e.g., the types of planning actions are known, as well as the structure of the IU analysis. This allows specifying the structure of the relational database and designing SQL-tables beforehand and reusing one database model (see Section~\ref{SSEC:RD}) in all deployments of the system. The database technology is a standard method of storing data, and checking validation (2) of the data is highly supported. This includes both simple checks of data types, as well as arbitrarily complex integrity checks with the use of database triggers. Once the database of a particular instance is populated, the designer can automatically generate a SNAP for a particular client/task/environment combination taking input for the sensors through the ubiquitous sensing technician's interface, and the POMDP can be fed into the planner, and then executed in the real environment or simulated once the planner computes the policy. Since, the overall process is straightforward for the designer, this allows for a traditional dynamic testing of the model, where the designer can adjust the domain model easily via the database interface, generate a new POMDP, and then simulate it and assess its prompting decisions. This shows that also maintenance (3) is well supported in our architecture. The SQL relational language is also flexible in representing structured (4) objects. In our work, it is used in conjunction with a workflow method (5), where the technology designer follows specific steps which require populating specific tables in the database. The relational database technology is one of the most popular ways of storing data, and it is vastly supported by tools (i.e., user friendly software/interfaces for entering and retrieving data) and those tools are nowadays becoming familiar even to a standard computer user. In our implementation, a PHP-based web interface is used, which from the designer's point of view does not differ from standard database-based systems. The formulation of our relational model as a database makes our design amechanistic which is a desirable property of software systems. By amechanistic, we mean the fact that the person performing the SNAP analysis and typing the outcome in our system operates on entities which correspond directly to the SNAP analysis and not necessarily to specific concepts of the POMDP model. This essentially means that knowledge of POMDPs is not required in order to complete model formulation.

Originally, the specification of POMDPs was performed by trained POMDP experts who would encode the problem in a specific notation that the planner can understand. The following definition of the POMDP expert is considered in this paper:
\begin{defin}
A {\em POMDP expert} is a person who knows the POMDP technology sufficiently well so that she can design, on her own, the POMDP planning domain and encode it in a formal language (either in a general purpose programming language or in a dedicated POMDP formalism).
\label{def_pomdp_expert}
\end{defin}
The person who does not have sufficient skills mentioned in Definition~\ref{def_pomdp_expert} will be referred to as a non-expert. The existing research investigated the problem of how the specification of POMDPs can be done by non-POMDP experts or people with limited knowledge about POMDPs, where the specialised tool helps designing the POMDP \citep{edelkamp05modplan,simpson07GIPO,vacquero05simple}. The key feature of such systems is that the designer still explicitly defines the planning problem. In this paper, we introduce another way of defining POMDPs through a translation of a psychological model. The designer is doing the psychological task analysis of the task, whereas the system takes the data provided by the designer and translates it into the valid POMDP automatically. This results in a new paradigm that frees the designer from the burden of knowing the POMDP technology.

\subsection{Probabilistic Relational Models\label{SSEC:PRM}}



In the application areas which are considered in this paper, planning problems are POMDPs. POMDPs can be seen as Dynamic Decision Networks (DDNs) \citep{RussellNorvig10}. In most POMDP planners, DDNs have a propositional representation, where the domain has a number of attributes, and attributes can take values from their corresponding domains. As long as this is satisfactory for planning purposes (i.e., when the POMDP has already been defined), the problem with designing methodologies and tools for engineering the definitions of such planning problems using propositional techniques is that the reuse of the model in new instances is not straightforward because the model ignores relations between various entities of the domain, and therefore a relational approach becomes useful. Statistical Relational Learning \citep{GetoorSRL07} makes relational specification of probabilistic models possible as it allows for defining different types of objects (classes) and their sets of attributes and relations between classes.   The specific model we focus on here is the probabilistic relational model, or PRM \citep{getoor07prm}.

\begin{figure}
  \centering
  \includegraphics[scale=0.8]{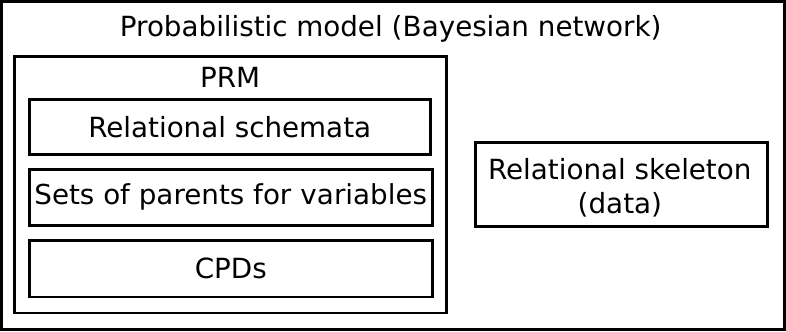} 
  \caption{The probabilistic model and its components when specified as a PRM. The relational schemata represents classes of objects, attributes/slots of classes, and relations between classes via reference slots. Sets of parents of variables specify which attributes influence other attributes in their conditional probability distributions. Conditional probability distributions are also required for every random variable. The relational skeleton provides specific objects for which the relational model allows obtaining a probability distribution over unobserved attributes of the skeleton.}
  \label{FIG:PRM}
\end{figure}
Probabilistic Relational Models (PRM) define a template for a probability distribution over attributes of objects (or columns of tables in database terminology) that specifies a concrete probability distribution when grounded with specific data.   Figure~\ref{FIG:PRM} shows the main components of the probabilistic model specified using the PRM. The first element is the relational schemata which can be formalised as a specification of types of objects (classes), their attributes, and relations between objects of specific types. The two additional components are: for each attribute the set of parents the attribute probabilistically depends on (and the aggregating method when the probability distribution depends on a variable number of objects), and the corresponding conditional probability distributions/tables (CPDs/CPTs).

A Bayesian network can be seen as a PRM with one class and no relations. The primary advantage of the PRM is that it distinguishes more than one class, implying that relationships between classes can be identified and then conditional probability tables (CPTs) can be shared among many objects. In particular, aggregating rules (details below) can facilitate CPTs which can depend on a changeable number of objects. Overall, models can be represented more compactly.

The relational specification of a POMDP also requires the reward function, which does not appear in the original definition of the PRM in \citep{getoor07prm}. Since any Bayesian network can be implicitly interpreted as a decision network when utilities are provided externally, or have explicit utility nodes \citep{Pearl88}, utilities are a natural extension to the original PRM.

The relational schemata of the PRM can be represented directly as a standard relational database where tables and derived tables determined by SQL queries define objects, and columns in tables define attributes. Relationships between objects are modelled as primary/foreign key constraints or by additional tables which model relationships/links. The key property of PRMs that we exploit in our model is that properties of objects can be represented either as attributes of classes/tables or as separate objects and connected with the main object using classes/tables which model relationships/links. Since conditional probability tables (CPTs) are parametrised, the CPT for a given type of attribute is the same in all possible objects, i.e., one CPT is reused for multiple objects. Aggregating functions of a variable number of objects are often used to achieve this property \citep{Pearl88,heckerman94anew_look,diez06}.

\section{Relational Database Model of SNAP\label{SSEC:RD}}
Our main emphasis is that a non-technical designer should be able to define the prompting system (the AI planning problem) easily and with minimal technical knowledge of probabilistic planning. The approach we are proposing in this paper is to provide a probabilistic relational schema  (as a relational database) which can be populated by the designer using standard database tools such as forms or web interfaces, and then grounding the PRM to a POMDP specification using a {\em generator} software, which implements parts of the PRM not stored in the database (such as aggregation operators).   In this section, we first give the PRM schema definition, followed by the precise specification of the aggregation operators that allow the probability distributions in the POMDP to be generated.

In the following, we use {\sf sans-serif} font for class names, {\em Capitalised Italics} for attributes and {\em lower-case italics} for objects or attribute values.  Set of attributes or values are in {\bf\em boldface italics}. The attributes, referred to also as slots, {\em A}, of a class, {\sf X} are denoted \slot{X}{A}.  A {\bf\em slot chain} is an attribute of a class that is an external reference to another class, and so consists of a set of objects, as \slotchainset{X}{A}.

\subsection{Relational Schemata}
The relational schema is composed of a set of classes, one for each variable (including observations and actions) in the POMDP shown in Figure~\ref{fig:pomdp}(b). The class name is the variable name.  The attributes/slots of each class include (at least)  an object {\attribute{Name}} (e.g. an ability object with {\em Name}= $rn\_cup\_on\_tray$) and a typed {\em Value}.  The value of each object is the value of the corresponding variable (or action) in the POMDP. We assume here without loss of generality that all variables except the {\em Task} are Boolean.  Any $N$-valued variable can always be converted to Boolean by splitting it into $N$ Boolean variables (thereby allowing concurrent values).  For example, in our previous work, we assume the client {\em Behaviour} is a single variable, and thus the client can only perform one behaviour (out of $N_B$ total) at a time. More generally, we can split this into $N_B$ Boolean variables and allow the client to be performing multiple behaviours simultaneously (e.g. talking on the phone and pouring water).  Similarly for system actions, we can allow concurrency or not.

In the following, we will examine the five different CPTs for the POMDP in Figure~\ref{fig:pomdp}(b), showing the relevant piece of the PRM, the dependency structure, and we will show how the CPT is derived using aggregation operators on the PRM. 

\subsubsection{Client's Abilities Dynamics Model - Y'} 
Actions of our prompting systems depend on specific abilities that the client has to posses in order to complete specific steps of the task. In our relational model, it is assumed that the system can either (a) prompt for a specific ability, or (b) do nothing (a no-op action that is present in all instantiations of the model).   The generator software creates one action for each ability in the domain. The prompt for an ability increases the chance that the client will regain that ability. For example, if the client lacks the ability to recognise the cup, the system may flash a light on the cup in order to focus client's attention on it.

\begin{figure}[bt]
  \begin{center}
\includegraphics[width=0.4\textwidth]{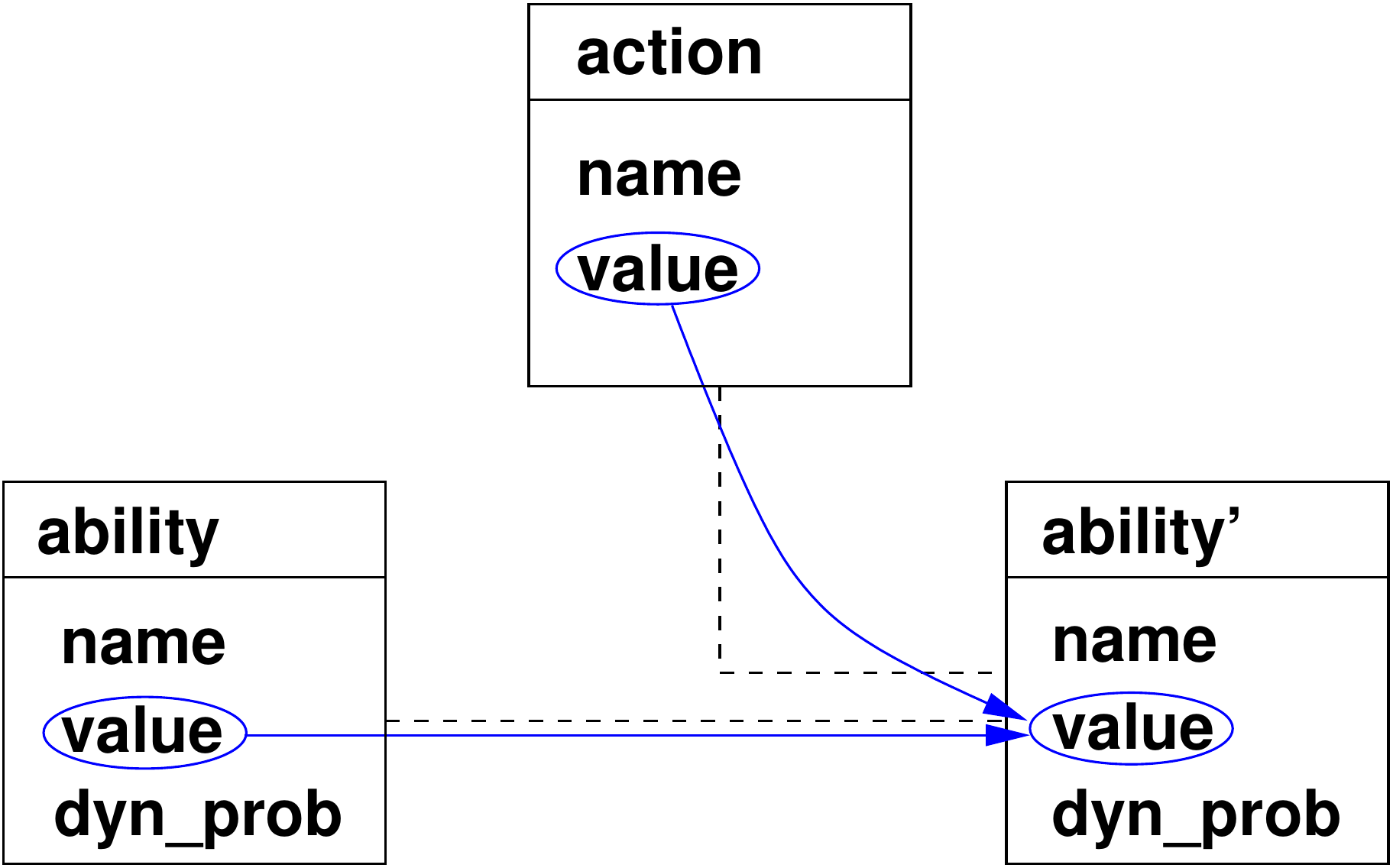}
\end{center}
\caption{\label{fig:prm-abilities} Relational schema and dependency structure for PRM for the client ability dynamics, $P(y'|y,a)$. Client's ability depends on the system's action and on the value of at least the same ability in the previous time step.}
\end{figure}

Figure~\ref{fig:prm-abilities} shows the dependency structure and relational schema for the PRM for the client's ability dynamics, $P(Y'|Y,A)$. 
The abilities have an additional slot {\bf\em Dyn\_prob} that includes a set of four probabilities: {\em \{keep\_prompt,gain\_prompt,keep,gain\}} that are used to define the CPT for the ability dynamics:
\[P(\text{\slotprime{Ability}{Value}}|\text{Pa(\slotprime{Ability}{Value})})\]
Here, the parents of the ability are
\begin{align*}
\text{Pa(\slotprime{Ability}{Value})} = \{&\gamma_1(\text{\slotprimechain{Ability}{Action}{Value}}),\\
&\gamma_2(\text{\slotprimechain{Ability}{Ability}{Value}})\}
\end{align*}
and $\gamma_1(\mathbf{A}) = \bigvee (\mathbf{A})$ is the aggregate (disjunction) of all actions that affect \slotprime{Ability}{Value}: any action can have this effect.
Similarly, $\gamma_2(\mathbf{Y})=\bigwedge (\mathbf{Y})$ is the aggregate (conjuction) all previous abilities that affect \slotprime{Ability}{Value} (usually this is only the same ability from the previous time-step).   
The CPT defines what happens to a particular client ability when an action is taken in a state where the client has a current set of abilities.  
If the action corresponds to that ability (e.g. is a prompt for it), then it will be in the set \slotprimechainset{Ability}{Action}, 
and thus $\gamma_1$ will be {\em true}.  The dynamics of an ability is also conditioned on the previous abilities the client had, 
as given by the set \slotprimechainset{Ability}{Ability}. If all these necessary precondition abilities are present, then $\gamma_2$ will be {\em true}.  
The CPT is then defined as follows, where we use the shorthand $Y'$ to denote \objectprime{Ability}{Value}, and $\mathbf{Y,A}$ to denote \objectslotprime{Ability}{Ability}, \objectslotprime{Ability}{Action}.

\begin{center}
\begin{tabular}{|c|c|l|}
\hline
$\gamma_1(\mathbf{A})$ & $\gamma_2(\mathbf{Y})$ & $P(Y'=true|\mathbf{Y,A})$ \\ \hline
true & true & \objectprime{Ability}{keep\_prompt} \\
true & false & \objectprime{Ability}{gain\_prompt}\\
false & true & \objectprime{Ability}{keep} \\
false & false & \objectprime{Ability}{gain}\\ \hline
\end{tabular}
\end{center}

Suppose a prompt is given for {\em ability$'$}. Then, if the ability preconditions are satisfied, the client will have the {\em ability} in the next time step with probability \objectprime{Ability}{keep\_prompt}.  If, on the other hand, the ability preconditions are not satisfied, this probability will be \objectprime{Ability}{gain\_prompt}.  The probabilities \objectprime{Ability}{keep} and \objectprime{Ability}{gain} are for the situation when the action is {\em not} related to the ability in question.

\subsubsection{Client Behaviour Dynamics Model - B'} 
\begin{figure}
  \begin{center}
\includegraphics[width=0.6\textwidth]{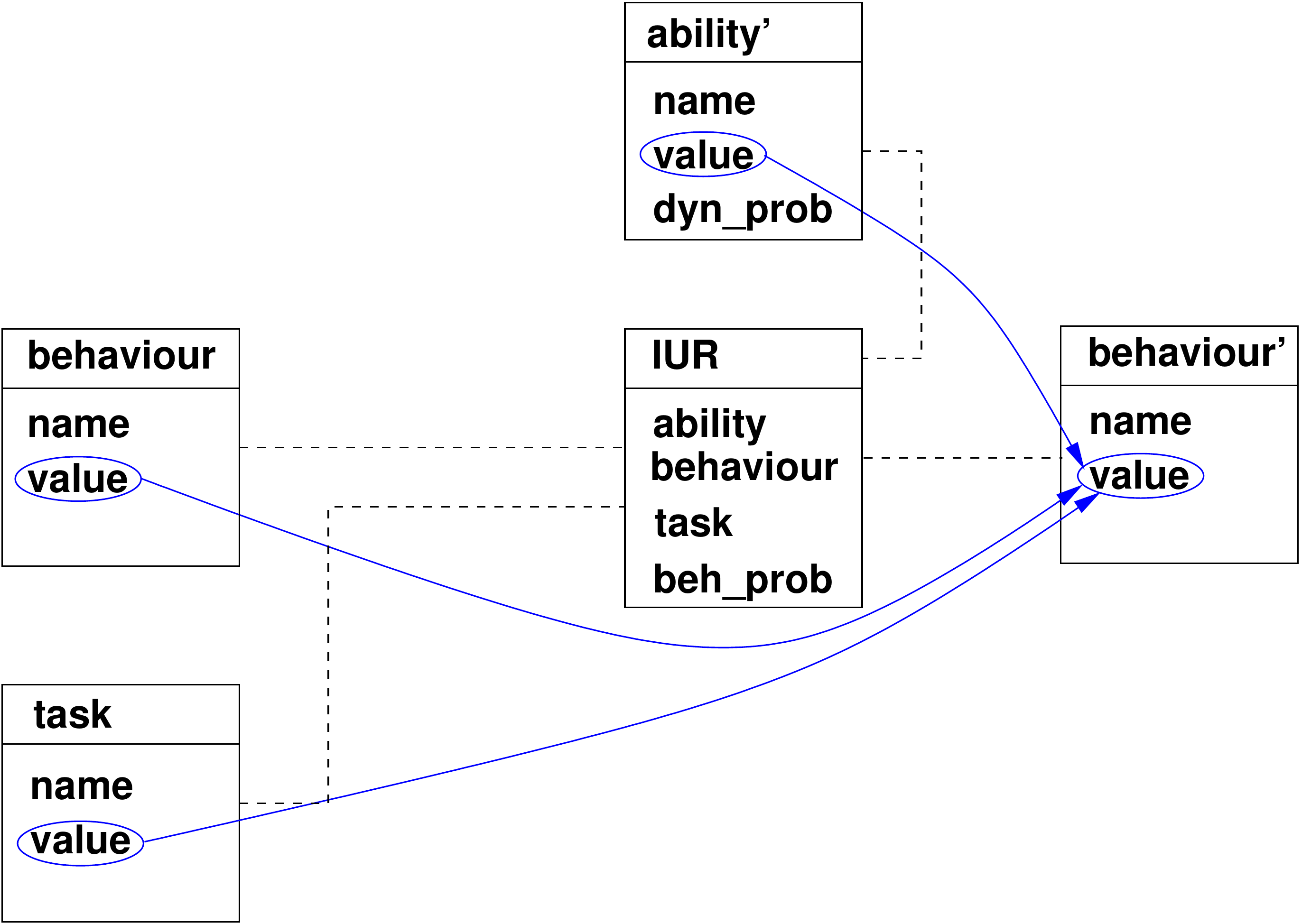}
\end{center}
\caption{\label{fig:prm-behaviours} Relational schema and dependency structure for PRM for the client behaviour dynamics, $P(B'|B,Y',T)$. The client's behaviour depends on the previous behaviour, current ability and the previous task state - this dependency originates from Table~\ref{tab:iuanal}.}
\end{figure}

Figure~\ref{fig:prm-behaviours} shows the dependency structure and relational schema for the PRM for the client behaviour dynamics, $P(B'|B,T,Y')$ (we leave off dependency on actions for clarity of presentation). 
In order to define the aggregation operators, we need to introduce a {\em link} class, {\sf\iurow}, that represents the IU table (e.g. Figure~\ref{tab:iuanal}).  An {\iurow} object is a single row of the IU table, and has sets of abilities, behaviours, and task variables, along with a {\em Beh\_prob} slot that gives the probability of a particular behaviour set occurring.  We are seeking the CPT:
\[P(\text{\slotprime{Behavior}{Value}}|Pa(\text{\slotprime{Behaviour}{Value}}))\]
where the parents of \slotprime{Behaviour}{Value} are an aggregate of \slotprimechain{Behaviour}{\iurow}{{\bf \em Ability$'$}}, \slotprimechain{Behaviour}{\iurow}{{\bf\em Task}} and \slotprimechain{Behaviour}{\iurow}{\bf\em Behaviour}.   However, we cannot define a variable as the aggregate, and instead directly define the probability distribution of interest.
\commentout{
\begin{align*}
\text{Pa(\slotprime{Behaviour}{value})} = \gamma(&\text{\slotprimechain{Behaviour}{\iurow}{Ability$'$}},\\
&\text{\slotprimechain{Behaviour}{\iurow}{Task}},\\
&\text{\slotprimechain{Behaviour}{\iurow}{Behaviour}})
\end{align*}
}

Denote the set of rows in the IU table as $I$. $T$, $T'$, $B$, $B'$, $Y$, and $Y'$ are as specified in Figure~\ref{fig:pomdp}(b). In order to compactly specify the CPT, we define the following Boolean aggregation functions:
\begin{enumerate}
  \item $row\_rel: I\times T \rightarrow \{0,1\}$ is 1 for task states relevant in row, $i$, and 0 otherwise, and aggregates all task variables in \slotprimechain{Behaviour}{\iurow}{\bf\em Task} as
\begin{equation}
row\_rel(i)=\bigwedge (\text{\slotprime{Behaviour}{row\_i.{\bf\em Task}.{\em Value}}}).
\label{eqn:bagg1}
\end{equation}
We write this as a function of the row, $i$, only: $row\_rel(i)$ leaving the remaining variables implicit. The same shorthand is applied to the other functions. 

 \item $row\_abil\_rel: I\times Y' \rightarrow \{0,1\}$ is 1 if all abilities on row $i$ are present, and aggregates all ability variables in \slotprimechain{Behaviour}{\iurow}{\bf\em Ability'}  as 
\begin{equation}
row\_abil\_rel(i) = \bigwedge( \text{\slotprime{Behaviour}{row\_i.{\bf\em Ability$'$\!}.{\em Value}}}).
\label{eqn:bagg2}
\end{equation}
\end{enumerate}
There may be multiple rows in the IU table with the same set of task states (i.e. there are two possible behaviours, possibly requiring different abilities, that are possible and relevant in a task state).  The {\iurow} class has an additional slot {\em beh\_prob(i)} that gives the probability that the behaviour on row $i$ will take place.  {\em beh\_prob(i)} must be a well defined probability distribution for each row, so if $behaviour(i)$ means the behaviour on row $i$, we must have
\begin{equation}\forall_b (\sum_{i| \text{behaviour(i)}=b} \text{\em beh\_prob(i)} = 1)\label{eqn:behprob}\end{equation}
With the two aggregations defined above and this probability, we can define the CPT of a behaviour, $B'$, if the set of rows on which that behaviour appears is $I_b$, as
\begin{equation}
P(B'=true|Y',T) = \bigvee_{i\in I_b} \left( \begin{array}{l} row\_abil\_rel(i)\land \\ row\_rel(i)\land \\ beh\_prob(i)\end{array}\right)
\label{eqn:pb}
\end{equation}
This distribution is formed as a disjunction of the conjunction of aggregations because there is some non-zero probability of the behaviour for {\em any} task state that appears in $row\_rel(i)$ where $i\in I_b$, and if the rows are not mutually exclusive, then the {\em beh\_prob(i)} should satisfy the condition (\ref{eqn:behprob}).  To evaluate this distribution, one takes some particular values for $Y',T = y',t$ and computes the aggregates given by 
(\ref{eqn:bagg1}) and (\ref{eqn:bagg2}). We have assumed that behaviours are Boolean (they are either present or not), but Equation~(\ref{eqn:pb}) can be extended to handle non-Boolean behaviours by also taking the conjunction with \slotprimechain{Behaviour}{\iurow}{Behaviour$'$.value}.


There are two special behaviours that does not appear in the IU table at all: {\em DoNothing} and {\em Other}.  {\em DoNothing} occurs either when no abilities are present for the current task, or when the state calls for doing nothing (a goal state or a state that does not appear in the IU table).   The probability distribution over the {\em DoNothing} behaviour can be formulated using the same aggregations as for the other behaviours, negated:
\begin{align}
P(&DoNothing=true|Y,T)\nonumber \\
&=\sum_{i\in I} [\lnot row\_abil\_rel(i)\land row\_rel(i)] \lor \\
&\prod_{i\in I}[ \lnot row\_rel(i) ]  \lor goal
\end{align}
where $goal$ specifies the set of goal states. 

Finally,  to ensure stability (no zero probabilities), we add a small constant $\rho$ to this distribution for any behaviour that is possible, and a second small weight, $\kappa$, on the behaviour if it was also present in the previous state.  In our current implementation, $\rho = 0.01$ and $\kappa=1$.

\begin{figure}[bt]
\begin{center}
\includegraphics[scale=0.4]{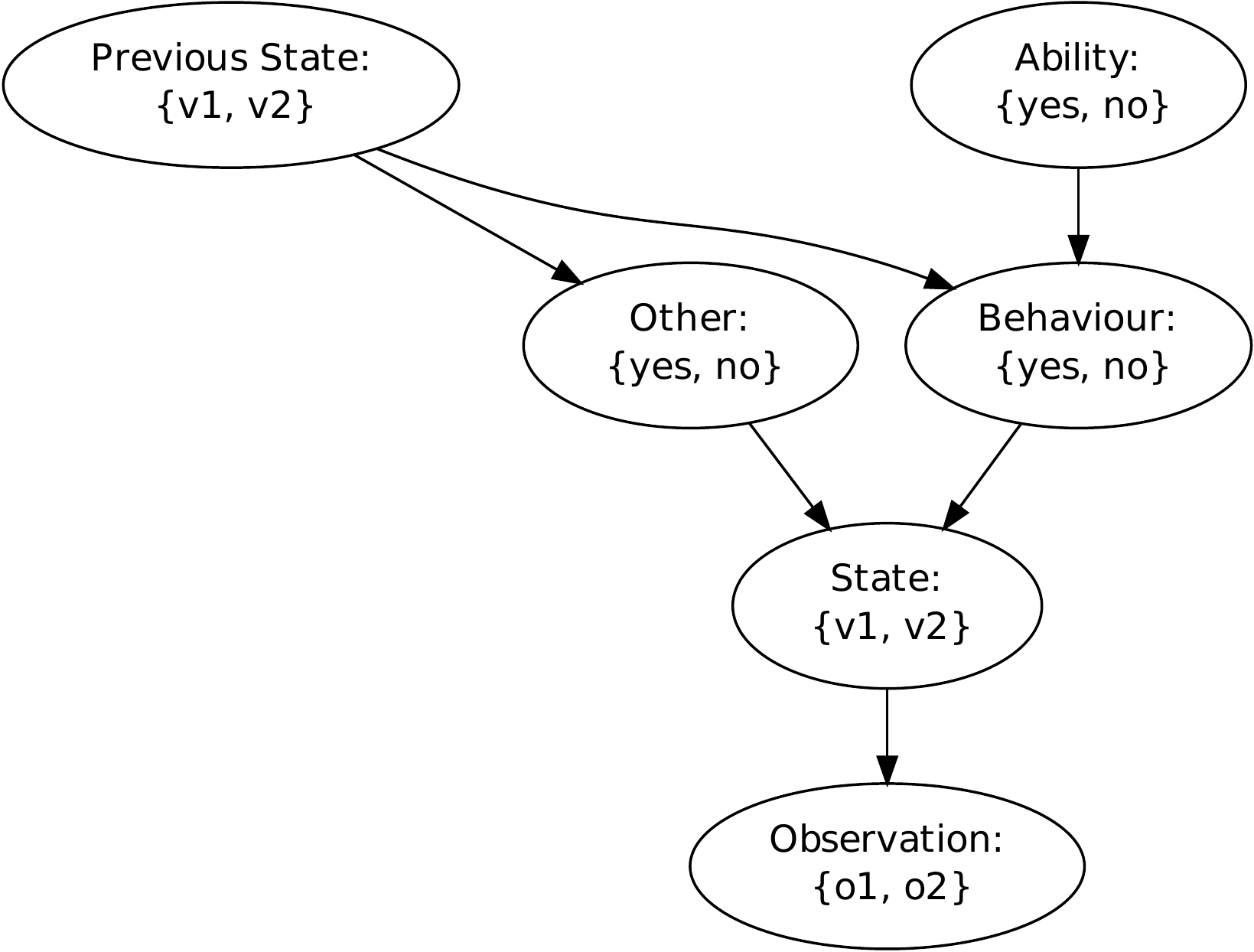}
\end{center}
\caption{A simple Bayesian network that explains how regression is implemented in our current system. Behaviour \texttt{other} yields a state transition to all values of the state with a strong preference towards keeping all features with their existing values.}
\label{fig_other_our}
\end{figure}

The behaviour {\em Other} handles regressions by the client in the task.  An assumption made by the SNAP methodology~\citep{SnapPMC11} is that the only reason a client will not do an expected behaviour is because of a lack of the required abilities.  Therefore, client behaviours are ``atomic'' by definition in the sense that the model cannot cope with multiple behaviours happening at once.  
However, the client can perform behaviours that cause a regression in the plan (e.g., when the client has finished washing her hands, she may take the soap again by mistake). 
In order to explain how our current implementation copes with regression, we use the example Bayesian network shown in Figure~\ref{fig_other_our}. This network shows generic nodes for the task states for two time steps: {\em previous state} and {\em state}, the ability, behaviour and observation. The problem of regression and non-atomic behaviours is mitigated by an additional artificial (non-existent in the IU analysis) behaviour {\em Other}. This behaviour does not require any abilities, and yields a state transition to all values of the state with a strong preference towards keeping all features with their existing values. With such a definition of behaviour {\em Other}, our system can adjust its belief state when the sensor readings provide information that contradicts the specified atomic behaviours, thereby alleviating the need for the designer to specify each and every possible regression.

\subsubsection{Task State Dynamics Model - T'} 
\begin{figure}
  \begin{center}
\includegraphics[width=0.6\textwidth]{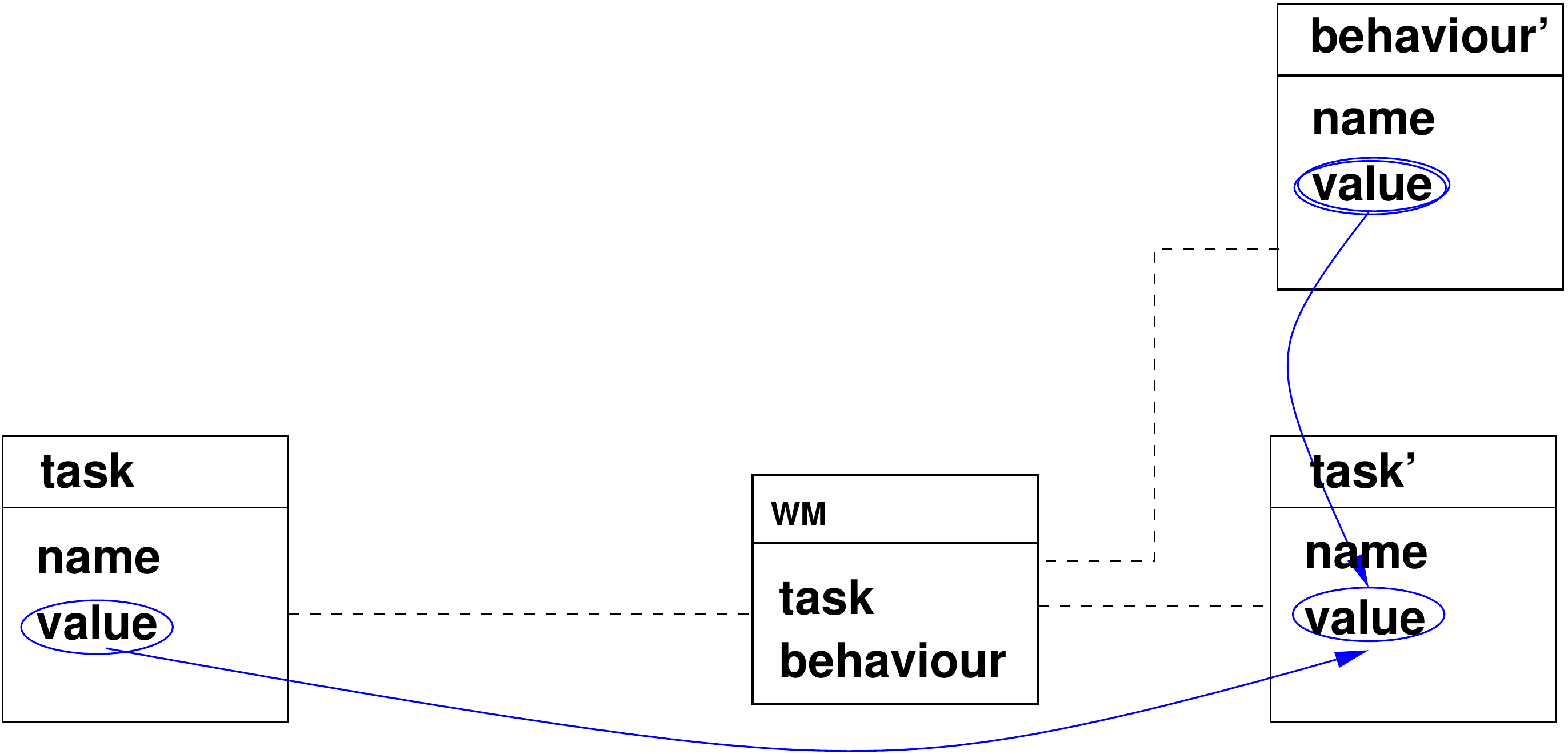}
\end{center}
\caption{\label{fig:prm-task} Relational schema and dependency structure for PRM for the client task dynamics, $P(t'|b',t)$. The state of the environment depends on its previous state and on the current behaviour of the client.}
\end{figure}

Figure~\ref{fig:prm-task} shows the dependency structure and relational schema for the PRM for the client task dynamics, $P(T'|B',T)$ 
(we leave off dependency on actions for clarity of presentation). 
In order to define the aggregation operators, we need to introduce a {\em link} class, {\sf WM}, that represents the world model describing how client behaviours change the state of the world.  A {\em WM} object has a single behaviour and a set of task variables (a state).  We are seeking the CPT:
\[P(\text{\slotprime{Task}{Value}}|Pa(\text{\slotprime{Task}{Value}}))\]
and this is defined over an aggregate of the parents of \slotprime{Task}{Value},  \slotprimechain{Task}{WM}{{\em Behaviour$'$}} and \slotprimechain{Task}{WM}{{\bf\em Task}} as follows:
\begin{equation}
P(T'|T,B') = \!\!\!\!\!\!\!\!\!\!\bigvee_{\begin{array}{l}wm_j \in \\\text{Task$'$\!.WM} \end{array}}\!\!\!\!\!\!\! \left(\begin{array}{l}\text{\slotchain{$wm_j$}{Task$'$\!}{Value}} \wedge \\ 
\text{\slot{$wm_j$}{Behaviour$'$\!}.{Value}} \wedge \\
\left[\bigwedge \text{\slotchain{$wm_j$}{Task}{Value}}\right] \end{array}\right)
\label{eqn:prm-task}
\end{equation}
For each task variable value, $t'$, this function is a disjunction over all behaviours, $b'$ that have $t'$ as an effect, and each term in the disjunction is a conjunction over all task variables that are preconditions for $b'$, and further includes $b'$ itself (\slot{$wm_j$}{Behaviour$'$\!}.{Value}) and the relevant value for $t'$ (\slotchain{$wm_j$}{Task}{Value}   - only necessary if the task variable in question is non-Boolean).

\subsubsection{Sensor Dynamics Model - K' and V'} For convenience, observations (sensors) are divided into states of sensors relevant to the task environment, K', and states of sensors relevant to client behaviour, V'.  We assume that each of the sensors depends on one (state or behaviour) variable only, and the sensor noises are represented explicitly in a table for each sensor and task variable combination. 

\subsubsection{Reward Model - R} The reward function is defined in a table that has a set of states and a reward value on each row.  Action costs are specified relative to each ability the action is meant to help with, in the abilities tables.


\commentout{
The following more complex SQL query example for $goal$ returns sets of states with the highest reward:
{\scriptsize
\begin{verbatim}
SELECT var_name, var_value, reward_value,
       t_rewards.state_set_id
FROM t_rewards_desc INNER JOIN t_rewards
ON t_rewards_desc.state_set_id=t_rewards.state_set_id
WHERE reward_value=(SELECT MAX(reward_value)
                FROM t_rewards)
ORDER BY 4  
\end{verbatim}}
}
\subsection{Implementation Details}
\begin{figure}
  \centering
  \includegraphics[scale=0.6]{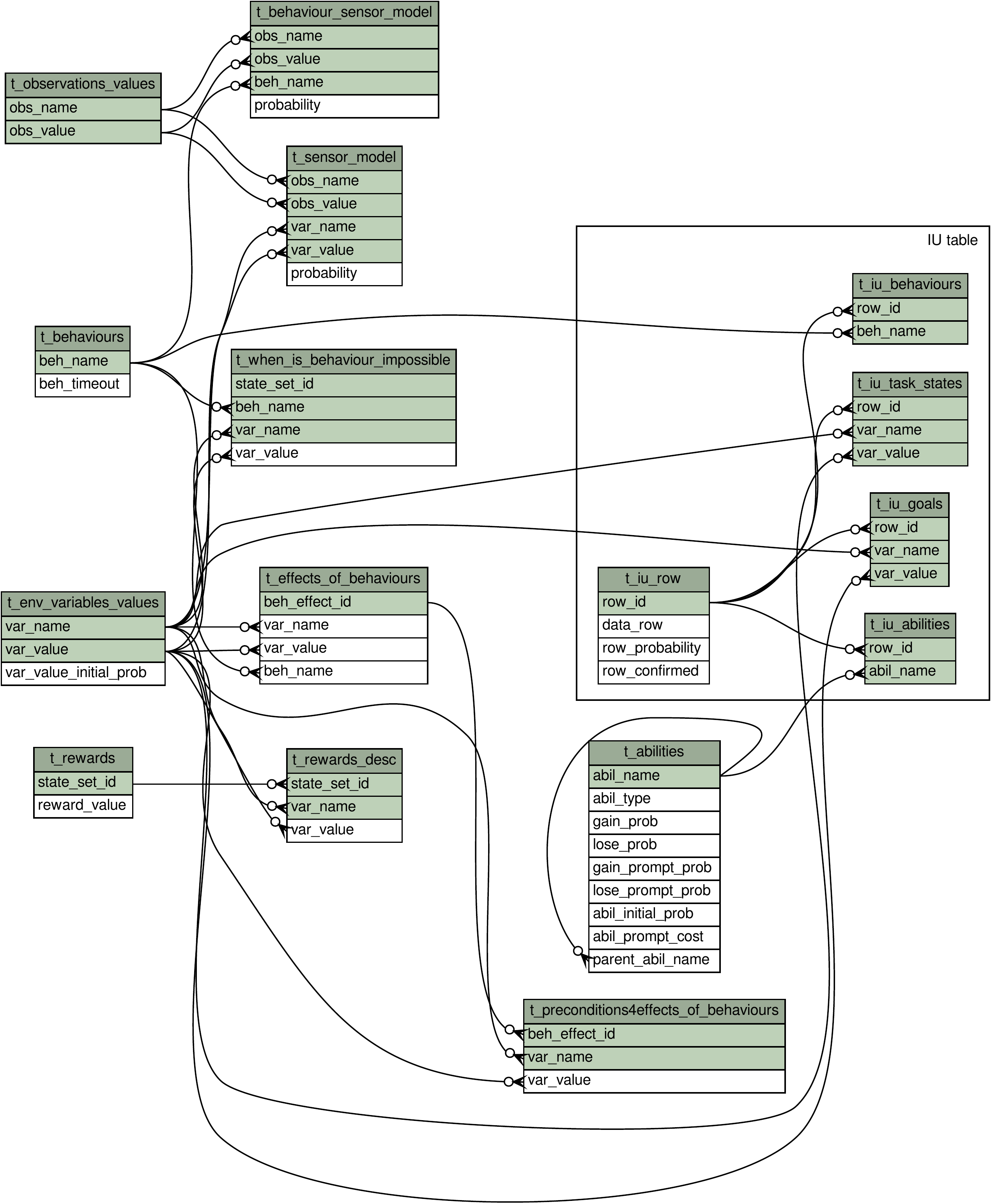}\\
  \caption{\label{FIG:DB}The diagram represents all tables which were implemented in the SQL database in order to store our SNAP model. Reference slots are indicated with lines connecting corresponding tables.}
  
\end{figure}

Figure~\ref{FIG:DB} shows the structure of the entire database that stores our relational schemata, probabilistic dependencies, and information required to define conditional probability tables. Note that the database does not explicitly distinguish between primed and unprimed variables, i.e., it defines variables once, and then the interpretation of the model as a dynamic Bayesian network repeats every variable twice for two times slices.
All tables that have their names starting with \texttt{t\_iu\_} represent the IU table.
The IU table view can be seen as a derived relation or table in the database: it is defined by an SQL join on the set of \texttt{t\_iu\_} tables. 
As shown in Figure~\ref{fig:prm-behaviours}, the rows of the IU table do not define random variables (there are no probabilistic links to attributes of the IU table in Figure~\ref{fig:prm-behaviours}), rather they are used in order to define conditional probability tables for \slotprime{Behaviour}{Value}. The task state attributes and their possible values are in table \texttt{t\_env\_vari\-ables\_val\-ues}. 
The sensors are defined in \texttt{t\_ob\-ser\-va\-tions\_val\-ues} and are linked with the behaviour via \texttt{t\_behaviour\_sensor\_model} or with the task via \texttt{t\_sensor\_model}. The last two tables are relationship classes that determine observations of the client and the task, and also define probabilities for sensor readings. The remaining tables that have \texttt{behaviour} in their name define dynamics of the client's behaviours. This includes effects and preconditions of behaviours, and also some additional constraints which specify when a behaviour is not possible. These behaviour related tables define what is shown in Figure~\ref{fig:prm-task} as the world model class. 
The preconditions and effects tables define the dynamics of behaviours that corresponds to what STRIPS operators \citep{fikes71strips} normally define in symbolic planning. 
Finally, rewards are defined in \texttt{t\_rewards} and the associated table allows specifying sets of states which yield a particular value of the reward.

\begin{figure}
  \centering
  \includegraphics[width=0.99\textwidth]{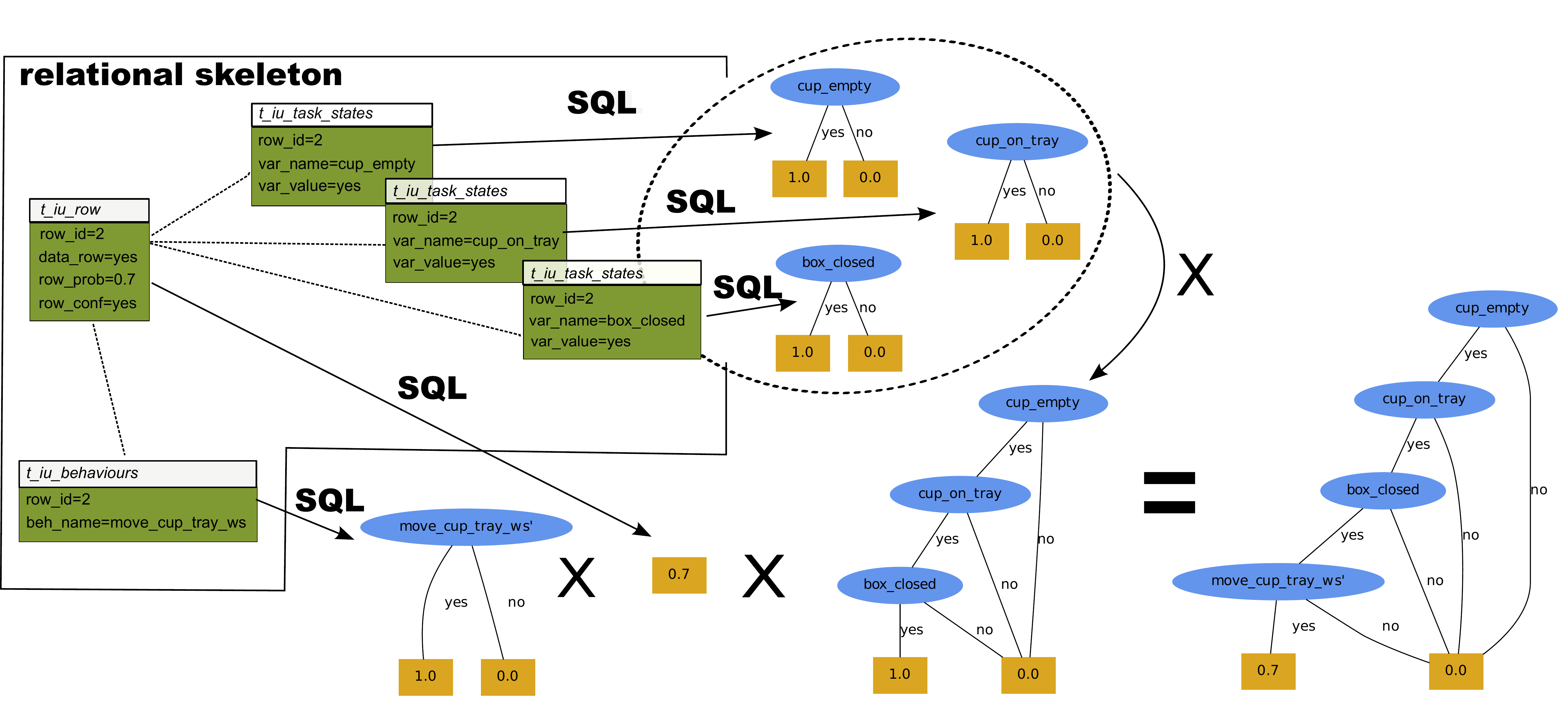}
  \caption{Subset of the relational skeleton for $P(b'|b,y,t')$ showing the generation of $behaviour(i,b')\wedge p(i)\wedge row\_rel(i) = row\_rel\_b(i,b') $ for row $i=2$ in Table~\ref{tab:iuanal}. The algebraic decision diagrams (ADDs) representing the relations are multiplied and normalised to give the final CPT.}
  \label{FIG:DDS}
\end{figure}
All the functions necessary to specify the POMDP are represented as algebraic decision diagrams (ADDs) in SPUDD notation~\citep{Hoey99}. These functions are computed with queries on the database. Each such query extracts some subset of ADDs from the relational skeleton. The ADDs are then combined using multiplication and addition to yield the final conditional probability tables (CPTs).
Some relations are explicitly represented in the database whereas others need to be extracted using more complex SQL queries. For example, data for $row\_rel(i)$, 
and $row\_abil\_rel(i)$ is read from the IU table in the database. The example subset of the relational skeleton for the IU analysis from Table~\ref{tab:iuanal}, and diagrams for selected functions are in Figure~\ref{FIG:DDS}.  SQL queries extract compound algebraic decision diagrams from the relational skeleton, and the generator software multiplies those diagrams in order to obtain final functions, such as $P(b'|b,y,t')$. 
A schematic is shown in Figure~\ref{FIG:DDS}, where the CPT for \objectprime{Behaviour}{move\_cup\_tray\_ws} is gathered from the relevant tables. The original table schema in the PRM for the relations in Figure~\ref{FIG:DDS} can be seen in Figure~\ref{FIG:DB}.

\subsection{Simulation and Validation} The database described in the
last section is provided as an online service. When the entire model
is specified in the database, the designer can generate a POMDP
automatically, and download a text-file specification of the
POMDP\footnote{for a complete demonstration, readers can view the video at \url{http://youtu.be/KHx9zGljkLY}, or alternatively can try out the system at \url{http://www.cs.uwaterloo.ca/~mgrzes/snap}}. We then provide software that solves the POMDP and allows for simulating it using a text-based input mechanism. Thus, the designer can experiment with how the model works in simulation, and in case of unexpected behaviour, the model can be adjusted and re-generated. This iterative process allows the designer to tune the reward function that would guarantee the desired behaviour of the system.  The simulation process is a key step, as it allows the designer, who has knowledge of the task domain and target clients but not of POMDPs, to see if her task knowledge corresponds to the model she has created.

\subsubsection{Constraints and Database Engines}

The advantage of the relational database is that it allows for easy implementation of the constraints required by the model. The simplest example are constraints on attribute values. For example, probabilities have to be in the range [0,~1], or literals should follow specific naming patterns (according to the requirements of the POMDP planner). These simple constraints are easily implemented in the definition of SQL tables. More complex constraints, which involve more than one attribute, are also required. For instance in the planner which we use, sensors and domain attributes are in the same name space, which means that their names have to be different. Furthermore, as we use directed graphical models, the acyclicity of our final DBN has to be guaranteed. Such things can be easily implemented using database triggers, and the designer will be prompted at the input time and informed about the constraint. The advantage of databases is that they allow for specifying and enforcing relations between different types of objects, for example, objects representing concepts in the environment and objects which serve as attributes of environment objects.

\subsubsection{Hierarchical Control}\label{sec:hier}

The IU analysis breaks an ADL like making a cup of tea down into a number of sub-tasks, or sub-goals. For tea making, there are five sub-goals. This decomposition arises naturally according to the major elements of recall noted in the videos from which this IU analysis was made. The five sub-goals are partially ordered, and the partial ordering can be specified as a list of pre-requisites for each sub-goal giving those sub-goals that must be completed prior to the sub-goal in question. Since each sub-goal is implemented as a separate POMDP controller, a mechanism is required to provide hi-level control to switch between sub-goals. We have implemented two such control mechanisms. A deterministic controller is described in~\citep{SnapPMC11}, and a probabilistic and hierarchical method in~\citep{Hoey11b}. The hierarchical control is however beyond the scope of this paper, because here our interest is in how the POMDP specification of one sub-task can be rapidly specified.

\section{Case Study with Encountered Pitfalls and Lessons Learned}\label{SEC:SYSUSE}

In this section, we provide a case study that explains how our tool is used in order to design a prompting system to help a person with dementia to wash their hands.  Our previous version of this system~\citep{Hoey10b} used a hand-crafted POMDP model, and our goal here is to show the process of creating a similar, but not identical, model using the PRM models and database introduced in the last section. 
 Normally, the IU analysis is completed in a spreadsheet~\citep{SnapPMC11}, and here we follow a designer, SC, going through the IU analysis using our database.  SC is an electrical engineer with no formal training in decision theory or Markov decision processes.  
As called for by the SNAP methodology~\citep{SnapPMC11}, SC started this process by watching multiple videos of clients washing their hands with the help of human caregivers. He then identified the elements as described in the following subsections.

\subsection{Task and Environment Features} 
\begin{figure}
  \centering
  \begin{tabular}{ccc}
  \includegraphics[scale=.8]{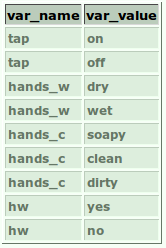} & \includegraphics[scale=.8]{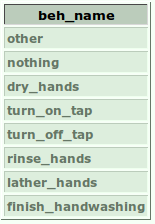} & \includegraphics[scale=.8]{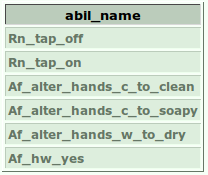}\\
  \texttt{t\_env\_variables\_values} & \texttt{t\_behaviours} & \texttt{t\_abilities} \\
  \end{tabular}
  \caption{The content of tables that define task states, behaviours and abilities in our hand washing example.}
  \label{FIG_CS_HWFEATURES}
\end{figure}

In the first step, SC identified the task states, client abilities and behaviours, and saved them in corresponding tables. The exact content of these tables (where some attributes are omitted for clarity) after performing this step is in Figure~\ref{FIG_CS_HWFEATURES}.  The designer is free to choose whatever names she pleases for the attributes.

\subsection{Dynamics of Client Behaviours} 
\begin{figure}
  \centering
  \begin{tabular}{ccc}
  \includegraphics[scale=.8]{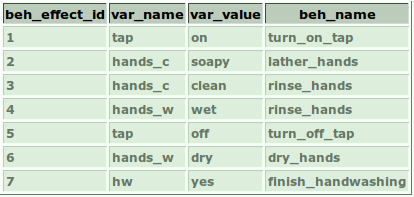} & \includegraphics[scale=.8]{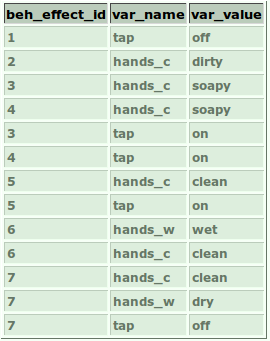} \\
  \texttt{t\_effects\_of\_behaviours} & \texttt{t\_preconditions4effects\_of\_behaviours} \\
  \end{tabular}
  \caption{Dynamics of behaviours in the hand washing example.}
  \label{FIG_CS_HWBEHDYN}
\end{figure}

Before specifying the IU table, SC specified details of behaviours of the client, where for every behaviour its preconditions and effects were specified in corresponding tables. The result of this step is in Figure~\ref{FIG_CS_HWBEHDYN}.
In these tables, every \textit{beh\_effect\_id} corresponds to one effect/precondition pair of a given behaviour. For example, rows with \textit{beh\_effect\_id=7} could be translated to the following STRIPS-like notation:
\begin{verbatim}
        Behaviour:     finish_handwashing
        Precondition:  hands_c = clean AND hands_w = dry AND tap = off  
        Effect:        hw = yes
\end{verbatim}

\subsection{IU Table} As in the original paper-based IU analysis, after completing the above two steps, SC created the IU table that represents the core of the design. For that, a number of interaction units were determined (the granularity depends on the dementia level of the client, and also on other client specific factors and preferences), and every interaction unit
corresponds to one row in the IU table. For every row, SC had to define the behaviour of the client, the task states that are relevant for that behaviour, and the abilities that are required for the behaviour to happen. It should be noted that the state relevance in the task state column usually makes stronger requirements about the behaviour than those specified in preconditions of the behaviour. This is because preconditions only say when the behaviour may be successful, whereas the IU table specifies when the behaviour will do something useful.
Before we show the exact IU table that SC created in this case study, we discuss several pitfalls encountered in this step, lessons learned and our solutions.

\subsubsection{State Subsumption in the IU Table}
\begin{figure}
  \centering
  \includegraphics[scale=0.8]{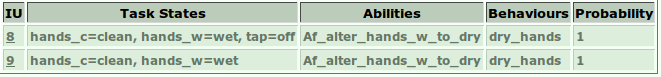}\\
  \caption{The example IU table where IU row 9 subsumes IU row 8. If the designer will not observe this, our system will do the check automatically.}
  \label{FIG_HWIUSUBSUMPTION}
\end{figure}

During the IU table specification, SC found that he needed to specify two interaction units with the same behaviour in the IU table that shared task states.
This indicates an error because both interaction units correspond the same behaviour, and one of them is the subset of the other. 
Figure~\ref{FIG_HWIUSUBSUMPTION} shows an example where IU row 9 subsumes IU row 8.
Our software indicates the presence of such dependencies and the designer needs to resolve them.  In this case, SC added the state feature \textit{tap=on} to row 9, resolving the issue.

\subsubsection{Overlapping State Relevance}
\begin{figure}
  \centering
  \includegraphics[scale=0.8]{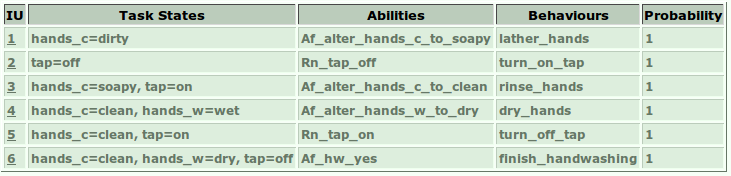}\\
  \caption{The initial IU table designed by the designer that contains behaviours that are have the same states relevant. For example, both \textit{lather\_hands} and \textit{turn\_on\_tap} have the same relevant state \textit{hands\_c=dirty, tap=off} that is not captured by the existing IU table.}
  \label{FIG_HWIUBEFOREEXPAN}
\end{figure}

\begin{figure}
  \centering
  \includegraphics[scale=0.8]{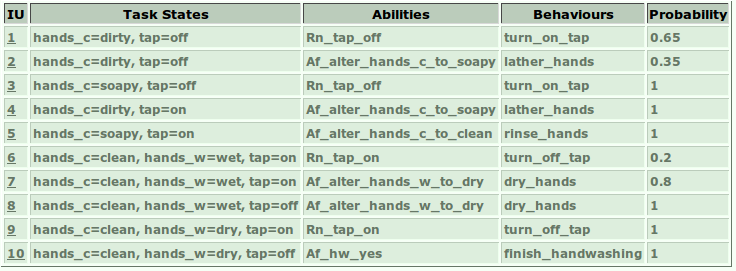}\\
  \caption{The IU table from Figure~\ref{FIG_HWIUBEFOREEXPAN} after automatic state expansion and removal of two generated rows that were undoing the progress towards the goal. Now, IU rows 1 and 2 are described by the same set of states that is relevant in two different behaviours and the likelihood of each is governed by the probability specified in column Probability.}
  \label{FIG_HWIUAFTEREXPAN}
\end{figure}

Sometimes there are multiple behaviours that can happen in a given state.  For example, in Figure~\ref{FIG_HWIUBEFOREEXPAN}, behaviours \textit{lather\_hands} and \textit{turn\_tap\_on} (from IU rows 1 and 2) have the same relevant state \textit{hands\_c=dirty, tap=off}.  However, this particular state is not explicitly stated in the IU table, since state relevance is defined on partially instantiated sets of states.
Although the designer correctly specifies two interaction units for the behaviours, she may not notice that these behaviours can happen in the same fully instantiated state.  
All states should be considered, states relevant for more than one behaviour identified, and states not relevant for any behaviour removed. For this reason, we introduced an automatic step in the process where the designer checks the subsumption of states and the overlapping state relevance.
This second check automatically adds more rows to the IU table, so that all sets of states 
relevant for all of the alternative behaviours are specified, for cases with overlapping state relevance. After this step, there may be several interaction units in the IU table that have the same set of relevant task states, but different behaviours. When this process is applied to the IU table in Figure~\ref{FIG_HWIUBEFOREEXPAN} the result is in Figure~\ref{FIG_HWIUAFTEREXPAN}  (row 1 has become rows 2 and 4, while row 2 has become rows 1 and 3). 
The result of the above transformation means that the client may do one of multiple behaviours, and the designer has to specify probabilities for each (column {\em Probability} in the IU table).
For example, the client may either turn on the tap or use the (pump) soap as a first step in handwashing (rows 1 and 2). Some clients may tend to start with turning the tap first, and the probability for that alternative would be higher as shown in Figure~\ref{FIG_HWIUAFTEREXPAN}. Similarly, in rows 6 and 7, we see that the client can turn off the tap or dry her hands in either order, but is more likely to dry hands first. 
The need for specifying this probability is one of the reasons why the expansion of the IU table cannot happen in the background when the system is generating the POMDP from the database and interaction with the designer is required. 

\vskip 2mm
In this case study, SC was very confident with the features of the system explained above and he found their use rather intuitive and straightforward. Figure~\ref{FIG_HWIUAFTEREXPAN} shows the final IU table that SC created in the hand washing task discussed in this case study. Note that the task states are defined in terms of partially instantiated states (sets of states) and the full instantiation of all states was not required.

\subsection{Rewards and Costs of Prompting} Once all the above steps are completed, the probabilistic dynamics of the core model are defined. The two missing things are rewards and sensors. The system allows the designer to specify the reward model explicitly in the database. We found that the following advice was sufficient for the non-POMDP professional, SC, to specify the reward model:
 \begin{itemize}
  \item 
SC was informed that the reward for finishing the task should be 15 (this number is chosen arbitrarily as the largest reward, but any linear scaling of reward functions is possible).
  \item Additionally a designer can specify an intermediate reward for reaching an intermediate goal. This reward should be smaller than the reward for achieving the final goal state, or can be negative if the domain has some dangerous situations that should be avoided (this does not happen in the handwashing task).
  \item The third element of the reward model is the cost of prompting. This specifies how costly it is to prompt for each ability. The value of this cost may be client dependent, because some clients may not like being asked specific questions. In general, recognition prompts are more invasive with this regard, and designers are advised to use higher costs for recognition prompts. 
 \end{itemize}

\begin{figure}
  \centering
  \includegraphics[scale=0.74]{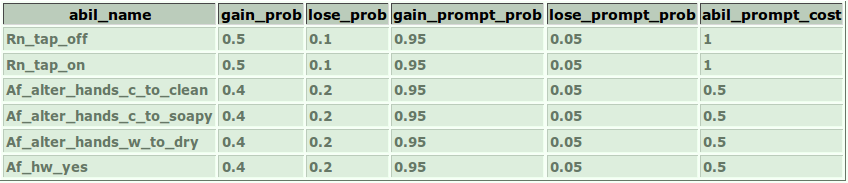}\\
  \caption{The example content of the \texttt{t\_abilities} table in one of our hand washing models. Ability \textit{Af\_hw\_yes} models the client being able to see the affordance of finishing the hand washing task. The cost 0.5 for this ability should be noted.}
  \label{FIG:HWABILITIESCOSTS}
\end{figure}

SC was following the above guidance in this case study and he defined the following rewards: R(\textit{hw=yes}) = 15 and R(\textit{hand\_c=clean}) = 3.
The initial costs of abilities that SC provided were as shown in Figure~\ref{FIG:HWABILITIESCOSTS}.
SC then noted that the resulting POMDP was issuing (in simulation) an unnecessary prompt for the ability \textit{Af\_hw\_yes}, i.e., after the client finished washing hands with \textit{hw=yes} the system would issue one more prompt for \textit{Af\_hw\_yes} that is required to make \textit{hw=yes}. This ability states that the client can see the affordance of finishing the task. The specific POMDP value function that SC used in the simulation (with original costs of prompts as specified in Figure~\ref{FIG:HWABILITIESCOSTS}) is as shown in Table~\ref{TAB:HWQVALUES}.
\begin{table}[bt]
\centering
\begin{tabular}{lll}
Action Name	&	Value of the Action	\\
\hline
donothing	&	350.349\\
prompt\_Af\_alter\_hands\_c\_to\_clean	&	349.795	\\
prompt\_Af\_alter\_hands\_c\_to\_soapy	&	349.727	\\
prompt\_Af\_alter\_hands\_w\_to\_dry	&	343.249	\\
prompt\_Af\_hw\_yes	&	350.456	\\
prompt\_Rn\_tap\_off	&	348.059	\\
prompt\_Rn\_tap\_on	&	349.295	\\
\end{tabular}  
\caption{The value of actions that prompt for abilities defined in Figure~\ref{FIG:HWABILITIESCOSTS} in the belief state that contains the marginal probability of variable hands washed, \textit{hw}, $P(hw=yes)=0.93$. Note that there is one prompt in this table for every ability from Figure~\ref{FIG:HWABILITIESCOSTS}. The action suggested by an optimal policy is in this case \textit{prompt\_Af\_hw\_yes}, i.e., an action with the highest value.}
\label{TAB:HWQVALUES}
\end{table} 
The value of the prompt \textit{Af\_hw\_yes} is slightly higher than the value of action {\em DoNothing}, but action {\em DoNothing} was the one that was expected by SC to happen in this particular situation. The exact full belief state is not printed here, but it had the marginal probability for \textit{hw=yes} being P(hw=yes)=0.93 that meant that the task has been completed with sufficiently high probability, and it was justified to require the {\em DoNothing} action from the system. 
When SC increased the cost  of prompting for \textit{Af\_hw\_yes} from 0.5 to 0.55, the system was correctly executing action {\em DoNothing} in such situations. This particular example shows how the designer can tune the reward function to suit his or her needs.
The designer can simply increase the cost, thus pushing the system to wait for more evidence that would justify a more expensive prompt. 
When the designer is running simulations for a given POMDP, the values of all system actions are displayed in a similar way as in Figure~\ref{TAB:HWQVALUES}.
This allows the designer to see the explanation for a specific behaviour and decide on the model change.

\subsection{Sensors} The last step is to specify the technical details that concern the physical deployment of the system in the environment, and our assumption is that these steps will be performed by hardware engineers. They will specify what kinds of sensors will be used, and they will provide the accuracy of those sensors that will be placed in tables that store the sensor model for behaviours and task states.

In the following section, we copmlete the case study with a demonstration of the resulting system in practice, along with simulated results for two other SNAP analyses.

\section{Demonstrative Examples\label{sec:experiments}}

We demonstrate the method on three assistance tasks: handwashing and toothbrushing with older adults with dementia, and on a factory assembly task for persons with a developmental disability. We show that once our relational system is designed (i.e. the database and the generator which reads the database and outputs the POMDP file), the system is generic and allows the designer to deploy the system in different tasks simply by populating the database for the new task.  The IU analysis for handwashing and toothbrushing were performed by a designer, SC, whose modelling work was reported in detail in our case study in Section~\ref{SEC:SYSUSE}. The analyses for the other two tasks were performed by a biomedical engineer, with limited experience with POMDPs or planning in AI.  As an example of the power of our method, the factory task was coded in about six hours by the engineer.  The factory task contained 6 different databases and associated POMDPs, each with about 5000 states, 24 observations, and 6 actions.
This can be compared to a manual coding of the system for handwashing (a smaller task), that took over 6 months of work resulting in the system described in~\citep{Boger05b}.

We clarify here that our experiments are not meant to demonstrate the final POMDP assistance systems working with real clients.  This is our eventual goal, but requires extensive additional work to set up clinical trials and recruit clients that is usually only started once working systems have been completely tested in simulation and laboratory environments.  In this paper, we are focussed on giving designers the ability to specify the models, and so we test the resulting models in simulations to check that they are consistent with the domains. We also test the hand washing model in the real life deployment of the system with a human actor who served as a potential client of the system. However, our SNAP analyses are performed using videos of real (potential) clients doing the actual tasks that we are designing for. 

\subsection{COACH and prompting}

Examples of automatic generation of task policies using IU analyses and a relational database were implemented for the task of handwashing and toothbrushing. For these examples, people with mild to moderate dementia living in a long term care facility were asked to wash their hands and brush their teeth in two separate trials. Videos were recorded and then used in order to derive the model of their corresponding tasks.

\subsubsection{Handwashing}

The washroom used to capture video for IU analysis for this task had a sink, pump-style soap dispenser and towel. Participants were led to the sink by a professional caregiver, and were encouraged to independently wash their own hands. The IU analysis was performed on eight videos captured from a camera mounted beside the sink. Each video was 2-4 minutes in length and involved one of six different clients with dementia (level not recorded). The task was broken into five steps: 1) turn on tap; 2) get soap; 3) rinse hands; 4) turn off water; and 5) dry hands. Steps 1 and 2 can be completed in any order, followed by step 3. After completion of step 3, steps 4 and 5 can be completed in any order. 
Figure~\ref{FIG_HWIUAFTEREXPAN} shows the IU table for the five steps and final virtual step (to provide an end condition), outlining the overall goals of the task, environmental state, and the client's abilities and actions relevant to the task. The goal hands\_c=clean (hands condition = clean) implies that the hands have been physically cleaned and rinsed, while the goal hw=yes (hands washed) indicates that the overall task is completed, which includes returning the environment to the original state.
\commentout{
\begin{table}
  \centering
  {\small
  \begin{tabular}{|l|l|l|l|l|}
\hline
IU & Goals & Task States & Abilities & Behaviours \\ \hline
1  & Hands Clean, Hands Washed  & hands\_c\_dirty           & Af: dispenser      & Lather hands \\ \hline
2  & Hands Clean, Hands Washed  & tap\_off                  & Rn: tap off        & Turn on tap \\ \hline
3  & Hands Clean, Hands Washed  & hands\_c\_soapy, tap\_on  & Af: water          & Rinse hands \\ \hline
4  & Hands Clean, Hands Washed  & hands\_c\_clean, hands\_w\_wet & Af: towel     & Dry hands \\ \hline
5  & Hands Clean, Hands Washed  & hands\_c\_clean, tap\_on  & Rn: tap on & Turn off tap \\ \hline
6  & Hands Clean, Hands Washed  & hands\_c\_clean, hands\_w\_dry & Af: hands washed & Task complete \\
   &              & tap\_off & & \\ \hline
  \end{tabular}
  }
  \caption{The IU analysis of the handwashing task.}
  \label{TAB:HANDIU}
\end{table} 
}

\subsubsection{Toothbrusing}

The videos used for the analysis captured participants in a washroom that had a sink, toothbrush, tube of toothpaste and cup, as they tried to independently brush their own teeth.  A formal caregiver was present to provide coaching and assistance if required. Six videos were used with six participants. The IU analysis was completed based on videos of several people and included multiple different methods of completing the task. The task was divided into 6 main steps, each containing multiple sub-steps: 1) wet brush; 2) apply toothpaste; 3) brush teeth; 4) clean mouth; 5) clean brush; and 6) tidy up. Steps 1 and 2 can be completed in any order, followed by step 3. Steps 4 and 5 can also be completed in any order after step 3, and step 6 is the final step following the first 5 steps.  
Table~\ref{TAB:TOOTHIU} shows the IU table for step 1 (wet brush), displaying the task and subtask goals, environmental state, and the client's abilities and actions relevant to the task.
\begin{table}
  \centering
  {\small
  \begin{tabular}{|l|l|l|l|l|}
\hline
IU & Goals & Task States & Abilities & Behaviours \\ \hline
1  & Final, wet brush & teeth\_dirty, tap\_off & Af: tap & Turn on tap \\ \hline
2  & Final, wet brush & teeth\_dirty, brush\_on\_surface & Rn: brush\_on\_surface & Take brush from surface \\ \hline
3  & Final, wet brush & teeth\_dirty, brush\_in\_cup & Rn: brush\_in\_cup & Take brush from cup \\ \hline
4  & Final, wet brush & teeth\_dirty, tap\_on, brush\_in\_hand & Af: water & Wet brush \\ \hline
  \end{tabular}
  }
  \caption{IU analysis of step 1 (wet brush) in the toothbrushing task. Step 1 in this task is about turning the water on, taking the tooth brush and making the brush wet.}
  \label{TAB:TOOTHIU}
\end{table} 

\subsubsection{Simulated Results on Hand Washing and Tooth Brushing}\label{S_HW_SIMUL}

A policy was generated for the handwashing task and for each sub-step of the toothbrushing task entered. Simulations were run to test the handwashing and toothbrushing policies 
by entering observations that corresponded to the client always forgetting steps of the task throughout the simulation (i.e., doing nothing) but responding to prompting if provided.  The simulations were run in a text-based interaction mode (no real sensors). Tables~\ref{TAB:HANDRESULTS} shows a sample of the belief state of the POMDP, the system's suggested action and the actual sensor states for several time steps during handwashing. Probabilities of the belief state are represented as the height of bars in corresponding columns of each time step. In the handwashing example, the client was  prompted to get soap (t=1, {\em Af\_alter\_hands\_c\_to\_soapy}). The client remained inactive (the same observations were received), so the system prompted the client to turn on the water (t=2, {\em Rn\_tap\_off}). The client turned on the water, so the system again prompted the client to get soap (t=3).  After detecting that the client had taken soap and the tap was on (preconditions for the next step) the system prompted the client to rinse off the soap (t=4,5, {\em Af\_alter\_hands\_c\_to\_clean}), which the client does at t=6. The client is then prompted to dry hands (t=6) and turn off the tap (t=7,8).

The toothbrushing simulation is shown in Table~\ref{TAB:TOOTHRESULTS}. At first, the system prompts the client to turn on the tap (t=1). The client does nothing, so the system tries to prompt the client to take the brush from the cup (in this case either turning the tap on or taking the toothbrush can happen first). The sensors indicate the brush was taken (t=3), so the system returns to prompting the client to turn on the tap.
\begin{table}[tbh]
  \setlength{\tabcolsep}{1pt}
  \centering
  \begin{tabular}{|r|cccc|ccccccccc|cccccccc|cccccc|c|}\hline
 & \multicolumn{4}{c|}{Observations}
 & \multicolumn{9}{c|}{Task}
 & \multicolumn{8}{c|}{Behaviour}
 & \multicolumn{6}{c|}{Ability}
 & \\ \cline{2-28}
 \begin{sideways}Step\end{sideways}
  & \begin{sideways}hands\_c\end{sideways}
  & \begin{sideways}hands\_w\end{sideways}
  & \begin{sideways}hw\end{sideways}
  & \begin{sideways}tap\end{sideways}
  & \begin{sideways}hands\_c\_clean\end{sideways}
  & \begin{sideways}hands\_c\_dirty\end{sideways}
  & \begin{sideways}hands\_c\_soapy\end{sideways}
  & \begin{sideways}hands\_w\_dry\end{sideways}
  & \begin{sideways}hands\_w\_wet\end{sideways}
  & \begin{sideways}hw\_no\end{sideways}
  & \begin{sideways}hw\_yes\end{sideways}
  & \begin{sideways}tap\_off\end{sideways}
  & \begin{sideways}tap\_on\end{sideways}
  & \begin{sideways}dry\_hands\end{sideways}
  & \begin{sideways}finish\_handwashing\end{sideways}
  & \begin{sideways}lather\_hands\end{sideways}
  & \begin{sideways}nothing\end{sideways}
  & \begin{sideways}other\end{sideways}
  & \begin{sideways}rinse\_hands\end{sideways}
  & \begin{sideways}turn\_off\_tap\end{sideways}
  & \begin{sideways}turn\_on\_tap\end{sideways}
  & \begin{sideways}Af\_alter\_hands\_c\_to\_clean\end{sideways}
  & \begin{sideways}Af\_alter\_hands\_c\_to\_soapy\end{sideways}
  & \begin{sideways}Af\_alter\_hands\_w\_to\_dry\end{sideways}
  & \begin{sideways}Af\_hw\_yes\end{sideways}
  & \begin{sideways}Rn\_tap\_off\end{sideways}
  & \begin{sideways}Rn\_tap\_on\end{sideways}
  & \begin{sideways}\makecell{System Action\\(prompt for ability specified)}\end{sideways}
\\ \hline
 0
 & dirty 
 & dry 
 & no 
 & off 
 & \rule{3mm}{0.01mm}
 & \rule{3mm}{2.00mm}
 & \rule{3mm}{0.01mm}
 & \rule{3mm}{2.00mm}
 & \rule{3mm}{0.01mm}
 & \rule{3mm}{2.00mm}
 & \rule{3mm}{0.01mm}
 & \rule{3mm}{2.00mm}
 & \rule{3mm}{0.01mm}
 & \rule{3mm}{0.01mm}
 & \rule{3mm}{0.01mm}
 & \rule{3mm}{0.01mm}
 & \rule{3mm}{0.01mm}
 & \rule{3mm}{0.01mm}
 & \rule{3mm}{0.01mm}
 & \rule{3mm}{0.01mm}
 & \rule{3mm}{0.01mm}
 & \rule{3mm}{1.60mm}
 & \rule{3mm}{1.60mm}
 & \rule{3mm}{1.60mm}
 & \rule{3mm}{1.60mm}
 & \rule{3mm}{1.90mm}
 & \rule{3mm}{1.90mm}
 & Af\_alter\_hands\_c\_to\_soapy 
\\ \hline
 1
 & dirty 
 & dry 
 & no 
 & off 
 & \rule{3mm}{0.01mm}
 & \rule{3mm}{1.94mm}
 & \rule{3mm}{0.06mm}
 & \rule{3mm}{2.00mm}
 & \rule{3mm}{0.01mm}
 & \rule{3mm}{2.00mm}
 & \rule{3mm}{0.01mm}
 & \rule{3mm}{1.92mm}
 & \rule{3mm}{0.08mm}
 & \rule{3mm}{0.28mm}
 & \rule{3mm}{0.28mm}
 & \rule{3mm}{0.04mm}
 & \rule{3mm}{0.56mm}
 & \rule{3mm}{0.22mm}
 & \rule{3mm}{0.28mm}
 & \rule{3mm}{0.28mm}
 & \rule{3mm}{0.06mm}
 & \rule{3mm}{1.44mm}
 & \rule{3mm}{1.84mm}
 & \rule{3mm}{1.44mm}
 & \rule{3mm}{1.44mm}
 & \rule{3mm}{1.60mm}
 & \rule{3mm}{1.76mm}
 & Af\_alter\_hands\_c\_to\_soapy 
\\ \hline
 2
 & dirty 
 & dry 
 & no 
 & off 
 & \rule{3mm}{0.01mm}
 & \rule{3mm}{1.96mm}
 & \rule{3mm}{0.04mm}
 & \rule{3mm}{2.00mm}
 & \rule{3mm}{0.01mm}
 & \rule{3mm}{2.00mm}
 & \rule{3mm}{0.01mm}
 & \rule{3mm}{1.94mm}
 & \rule{3mm}{0.06mm}
 & \rule{3mm}{0.24mm}
 & \rule{3mm}{0.24mm}
 & \rule{3mm}{0.04mm}
 & \rule{3mm}{0.78mm}
 & \rule{3mm}{0.16mm}
 & \rule{3mm}{0.24mm}
 & \rule{3mm}{0.24mm}
 & \rule{3mm}{0.06mm}
 & \rule{3mm}{1.38mm}
 & \rule{3mm}{1.86mm}
 & \rule{3mm}{1.38mm}
 & \rule{3mm}{1.38mm}
 & \rule{3mm}{1.48mm}
 & \rule{3mm}{1.70mm}
 & Rn\_tap\_off 
\\ \hline
 3
 & dirty 
 & dry 
 & no 
 & on 
 & \rule{3mm}{0.01mm}
 & \rule{3mm}{2.00mm}
 & \rule{3mm}{0.01mm}
 & \rule{3mm}{2.00mm}
 & \rule{3mm}{0.01mm}
 & \rule{3mm}{2.00mm}
 & \rule{3mm}{0.01mm}
 & \rule{3mm}{0.18mm}
 & \rule{3mm}{1.82mm}
 & \rule{3mm}{0.02mm}
 & \rule{3mm}{0.02mm}
 & \rule{3mm}{0.01mm}
 & \rule{3mm}{0.12mm}
 & \rule{3mm}{0.02mm}
 & \rule{3mm}{0.02mm}
 & \rule{3mm}{0.02mm}
 & \rule{3mm}{1.78mm}
 & \rule{3mm}{1.36mm}
 & \rule{3mm}{1.60mm}
 & \rule{3mm}{1.36mm}
 & \rule{3mm}{1.36mm}
 & \rule{3mm}{1.98mm}
 & \rule{3mm}{1.68mm}
 & Af\_alter\_hands\_c\_to\_soapy 
\\ \hline
 4
 & soapy 
 & dry 
 & no 
 & on 
 & \rule{3mm}{0.01mm}
 & \rule{3mm}{0.12mm}
 & \rule{3mm}{1.88mm}
 & \rule{3mm}{2.00mm}
 & \rule{3mm}{0.01mm}
 & \rule{3mm}{2.00mm}
 & \rule{3mm}{0.01mm}
 & \rule{3mm}{0.01mm}
 & \rule{3mm}{2.00mm}
 & \rule{3mm}{0.01mm}
 & \rule{3mm}{0.01mm}
 & \rule{3mm}{1.86mm}
 & \rule{3mm}{0.02mm}
 & \rule{3mm}{0.01mm}
 & \rule{3mm}{0.01mm}
 & \rule{3mm}{0.01mm}
 & \rule{3mm}{0.12mm}
 & \rule{3mm}{1.34mm}
 & \rule{3mm}{2.00mm}
 & \rule{3mm}{1.34mm}
 & \rule{3mm}{1.34mm}
 & \rule{3mm}{1.80mm}
 & \rule{3mm}{1.68mm}
 & Af\_alter\_hands\_c\_to\_clean 
\\ \hline
 5
 & soapy 
 & dry 
 & no 
 & on 
 & \rule{3mm}{0.01mm}
 & \rule{3mm}{0.01mm}
 & \rule{3mm}{2.00mm}
 & \rule{3mm}{2.00mm}
 & \rule{3mm}{0.01mm}
 & \rule{3mm}{2.00mm}
 & \rule{3mm}{0.01mm}
 & \rule{3mm}{0.01mm}
 & \rule{3mm}{2.00mm}
 & \rule{3mm}{0.02mm}
 & \rule{3mm}{0.01mm}
 & \rule{3mm}{1.80mm}
 & \rule{3mm}{0.10mm}
 & \rule{3mm}{0.02mm}
 & \rule{3mm}{0.01mm}
 & \rule{3mm}{0.02mm}
 & \rule{3mm}{0.04mm}
 & \rule{3mm}{1.82mm}
 & \rule{3mm}{1.62mm}
 & \rule{3mm}{1.34mm}
 & \rule{3mm}{1.34mm}
 & \rule{3mm}{1.72mm}
 & \rule{3mm}{1.66mm}
 & Af\_alter\_hands\_c\_to\_clean 
\\ \hline
 6
 & clean 
 & wet 
 & no 
 & on 
 & \rule{3mm}{2.00mm}
 & \rule{3mm}{0.01mm}
 & \rule{3mm}{0.01mm}
 & \rule{3mm}{0.01mm}
 & \rule{3mm}{2.00mm}
 & \rule{3mm}{2.00mm}
 & \rule{3mm}{0.01mm}
 & \rule{3mm}{0.01mm}
 & \rule{3mm}{2.00mm}
 & \rule{3mm}{0.01mm}
 & \rule{3mm}{0.01mm}
 & \rule{3mm}{0.01mm}
 & \rule{3mm}{0.01mm}
 & \rule{3mm}{0.01mm}
 & \rule{3mm}{2.00mm}
 & \rule{3mm}{0.01mm}
 & \rule{3mm}{0.01mm}
 & \rule{3mm}{2.00mm}
 & \rule{3mm}{1.44mm}
 & \rule{3mm}{1.34mm}
 & \rule{3mm}{1.34mm}
 & \rule{3mm}{1.68mm}
 & \rule{3mm}{1.66mm}
 & Af\_alter\_hands\_w\_to\_dry 
\\ \hline
 7
 & clean 
 & dry 
 & no 
 & on 
 & \rule{3mm}{2.00mm}
 & \rule{3mm}{0.01mm}
 & \rule{3mm}{0.01mm}
 & \rule{3mm}{1.50mm}
 & \rule{3mm}{0.50mm}
 & \rule{3mm}{2.00mm}
 & \rule{3mm}{0.01mm}
 & \rule{3mm}{0.02mm}
 & \rule{3mm}{1.98mm}
 & \rule{3mm}{1.50mm}
 & \rule{3mm}{0.01mm}
 & \rule{3mm}{0.01mm}
 & \rule{3mm}{0.08mm}
 & \rule{3mm}{0.01mm}
 & \rule{3mm}{0.40mm}
 & \rule{3mm}{0.02mm}
 & \rule{3mm}{0.01mm}
 & \rule{3mm}{1.60mm}
 & \rule{3mm}{1.38mm}
 & \rule{3mm}{1.96mm}
 & \rule{3mm}{1.34mm}
 & \rule{3mm}{1.68mm}
 & \rule{3mm}{1.64mm}
 & Rn\_tap\_on 
\\ \hline
 8
 & clean 
 & dry 
 & no 
 & on 
 & \rule{3mm}{2.00mm}
 & \rule{3mm}{0.01mm}
 & \rule{3mm}{0.01mm}
 & \rule{3mm}{1.96mm}
 & \rule{3mm}{0.04mm}
 & \rule{3mm}{2.00mm}
 & \rule{3mm}{0.01mm}
 & \rule{3mm}{0.08mm}
 & \rule{3mm}{1.92mm}
 & \rule{3mm}{1.70mm}
 & \rule{3mm}{0.02mm}
 & \rule{3mm}{0.02mm}
 & \rule{3mm}{0.10mm}
 & \rule{3mm}{0.02mm}
 & \rule{3mm}{0.04mm}
 & \rule{3mm}{0.08mm}
 & \rule{3mm}{0.02mm}
 & \rule{3mm}{1.44mm}
 & \rule{3mm}{1.36mm}
 & \rule{3mm}{1.62mm}
 & \rule{3mm}{1.34mm}
 & \rule{3mm}{1.68mm}
 & \rule{3mm}{1.82mm}
 & Rn\_tap\_on 
\\ \hline
 9
 & clean 
 & dry 
 & no 
 & off 
 & \rule{3mm}{2.00mm}
 & \rule{3mm}{0.01mm}
 & \rule{3mm}{0.01mm}
 & \rule{3mm}{2.00mm}
 & \rule{3mm}{0.01mm}
 & \rule{3mm}{2.00mm}
 & \rule{3mm}{0.01mm}
 & \rule{3mm}{1.90mm}
 & \rule{3mm}{0.10mm}
 & \rule{3mm}{0.10mm}
 & \rule{3mm}{0.01mm}
 & \rule{3mm}{0.01mm}
 & \rule{3mm}{0.04mm}
 & \rule{3mm}{0.01mm}
 & \rule{3mm}{0.01mm}
 & \rule{3mm}{1.86mm}
 & \rule{3mm}{0.01mm}
 & \rule{3mm}{1.38mm}
 & \rule{3mm}{1.34mm}
 & \rule{3mm}{1.44mm}
 & \rule{3mm}{1.32mm}
 & \rule{3mm}{1.66mm}
 & \rule{3mm}{2.00mm}
 & Af\_hw\_yes 
\\ \hline
 10
 & clean 
 & dry 
 & yes 
 & off 
 & \rule{3mm}{2.00mm}
 & \rule{3mm}{0.01mm}
 & \rule{3mm}{0.01mm}
 & \rule{3mm}{2.00mm}
 & \rule{3mm}{0.01mm}
 & \rule{3mm}{0.12mm}
 & \rule{3mm}{1.88mm}
 & \rule{3mm}{2.00mm}
 & \rule{3mm}{0.01mm}
 & \rule{3mm}{0.01mm}
 & \rule{3mm}{1.88mm}
 & \rule{3mm}{0.01mm}
 & \rule{3mm}{0.01mm}
 & \rule{3mm}{0.01mm}
 & \rule{3mm}{0.01mm}
 & \rule{3mm}{0.12mm}
 & \rule{3mm}{0.01mm}
 & \rule{3mm}{1.36mm}
 & \rule{3mm}{1.34mm}
 & \rule{3mm}{1.38mm}
 & \rule{3mm}{2.00mm}
 & \rule{3mm}{1.68mm}
 & \rule{3mm}{1.80mm}
 & donothing 
\\ \hline

  \end{tabular}
  \caption{Example simulation in the handwashing task. All simulations are presented in the same way, where every row represents one time step. The sensor readings are shown in the observations column. The current belief estimation is reflected by the height of the bar in the columns corresponding to specific random variables (task, behaviour and ability). The last column shows the prompt suggested by the system.}
  \label{TAB:HANDRESULTS}
\end{table}

\begin{table}[tbh]
  \setlength{\tabcolsep}{1pt}
  \centering
  \begin{tabular}{|r|ccc|ccccc|cccccc|cccc|c|}\hline
 & \multicolumn{3}{c|}{Observations}
 & \multicolumn{5}{c|}{Task}
 & \multicolumn{6}{c|}{Behaviour}
 & \multicolumn{4}{c|}{Ability}
 & \\ \cline{2-19}
 \begin{sideways}Time step, t\end{sideways}
  & \begin{sideways}brush\_wet\end{sideways}
  & \begin{sideways}tap\end{sideways}
  & \begin{sideways}brush\_position\end{sideways}
  & \begin{sideways}brush\_wet\end{sideways}
  & \begin{sideways}brush\_in\_hand\end{sideways}
  & \begin{sideways}brush\_in\_cup\end{sideways}
  & \begin{sideways}brush\_on\_surface\end{sideways}
  & \begin{sideways}tap\_on\end{sideways}
  & \begin{sideways}other\end{sideways}
  & \begin{sideways}nothing\end{sideways}
  & \begin{sideways}alter\_tap\_to\_on\end{sideways}
  & \begin{sideways}take\_brush\_from\_cup\end{sideways}
  & \begin{sideways}wet\_brush\end{sideways}
  & \begin{sideways}take\_brush\_from\_surface\end{sideways}
  & \begin{sideways}Rn\_brush\_in\_cup\end{sideways}
  & \begin{sideways}Rn\_brush\_on\_surface\end{sideways}
  & \begin{sideways}Af\_tap\end{sideways}
  & \begin{sideways}Af\_water\end{sideways}
  & \begin{sideways}\makecell{System Action\\(prompt for ability specified)}\end{sideways}
\\ \hline
 0
 & dry 
 & off 
 & in\_cup 
 & \rule{3mm}{0.01mm}
 & \rule{3mm}{0.20mm}
 & \rule{3mm}{1.20mm}
 & \rule{3mm}{0.60mm}
 & \rule{3mm}{0.01mm}
 & \rule{3mm}{0.01mm}
 & \rule{3mm}{0.01mm}
 & \rule{3mm}{0.01mm}
 & \rule{3mm}{0.01mm}
 & \rule{3mm}{0.01mm}
 & \rule{3mm}{0.01mm}
 & \rule{3mm}{1.90mm}
 & \rule{3mm}{1.90mm}
 & \rule{3mm}{1.60mm}
 & \rule{3mm}{1.60mm}
 & Af\_tap 
\\ \hline
 1
 & dry 
 & off 
 & in\_cup 
 & \rule{3mm}{0.01mm}
 & \rule{3mm}{0.14mm}
 & \rule{3mm}{1.82mm}
 & \rule{3mm}{0.06mm}
 & \rule{3mm}{0.08mm}
 & \rule{3mm}{0.40mm}
 & \rule{3mm}{0.64mm}
 & \rule{3mm}{0.08mm}
 & \rule{3mm}{0.08mm}
 & \rule{3mm}{0.40mm}
 & \rule{3mm}{0.40mm}
 & \rule{3mm}{1.62mm}
 & \rule{3mm}{1.76mm}
 & \rule{3mm}{1.96mm}
 & \rule{3mm}{1.44mm}
 & Af\_tap 
\\ \hline
 2
 & dry 
 & off 
 & in\_cup 
 & \rule{3mm}{0.01mm}
 & \rule{3mm}{0.04mm}
 & \rule{3mm}{1.96mm}
 & \rule{3mm}{0.01mm}
 & \rule{3mm}{0.04mm}
 & \rule{3mm}{0.36mm}
 & \rule{3mm}{0.82mm}
 & \rule{3mm}{0.04mm}
 & \rule{3mm}{0.04mm}
 & \rule{3mm}{0.36mm}
 & \rule{3mm}{0.36mm}
 & \rule{3mm}{1.50mm}
 & \rule{3mm}{1.70mm}
 & \rule{3mm}{2.00mm}
 & \rule{3mm}{1.38mm}
 & Rn\_brush\_cup 
\\ \hline
 3
 & dry 
 & off 
 & in\_hand 
 & \rule{3mm}{0.01mm}
 & \rule{3mm}{1.78mm}
 & \rule{3mm}{0.22mm}
 & \rule{3mm}{0.01mm}
 & \rule{3mm}{0.02mm}
 & \rule{3mm}{0.04mm}
 & \rule{3mm}{0.14mm}
 & \rule{3mm}{0.01mm}
 & \rule{3mm}{1.74mm}
 & \rule{3mm}{0.04mm}
 & \rule{3mm}{0.04mm}
 & \rule{3mm}{2.00mm}
 & \rule{3mm}{1.68mm}
 & \rule{3mm}{1.62mm}
 & \rule{3mm}{1.36mm}
 & Af\_tap 
\\ \hline
 4
 & dry 
 & on 
 & in\_hand 
 & \rule{3mm}{0.01mm}
 & \rule{3mm}{2.00mm}
 & \rule{3mm}{0.01mm}
 & \rule{3mm}{0.01mm}
 & \rule{3mm}{1.88mm}
 & \rule{3mm}{0.01mm}
 & \rule{3mm}{0.01mm}
 & \rule{3mm}{1.88mm}
 & \rule{3mm}{0.12mm}
 & \rule{3mm}{0.01mm}
 & \rule{3mm}{0.01mm}
 & \rule{3mm}{1.80mm}
 & \rule{3mm}{1.68mm}
 & \rule{3mm}{2.00mm}
 & \rule{3mm}{1.34mm}
 & Af\_water 
\\ \hline
 5
 & dry 
 & on 
 & in\_hand 
 & \rule{3mm}{0.08mm}
 & \rule{3mm}{2.00mm}
 & \rule{3mm}{0.01mm}
 & \rule{3mm}{0.01mm}
 & \rule{3mm}{2.00mm}
 & \rule{3mm}{0.01mm}
 & \rule{3mm}{0.06mm}
 & \rule{3mm}{1.80mm}
 & \rule{3mm}{0.04mm}
 & \rule{3mm}{0.10mm}
 & \rule{3mm}{0.01mm}
 & \rule{3mm}{1.72mm}
 & \rule{3mm}{1.68mm}
 & \rule{3mm}{1.62mm}
 & \rule{3mm}{1.94mm}
 & Af\_water 
\\ \hline
 6
 & wet 
 & on 
 & in\_hand 
 & \rule{3mm}{1.90mm}
 & \rule{3mm}{2.00mm}
 & \rule{3mm}{0.01mm}
 & \rule{3mm}{0.01mm}
 & \rule{3mm}{2.00mm}
 & \rule{3mm}{0.01mm}
 & \rule{3mm}{0.06mm}
 & \rule{3mm}{0.08mm}
 & \rule{3mm}{0.01mm}
 & \rule{3mm}{1.84mm}
 & \rule{3mm}{0.01mm}
 & \rule{3mm}{1.68mm}
 & \rule{3mm}{1.66mm}
 & \rule{3mm}{1.44mm}
 & \rule{3mm}{2.00mm}
 & donothing 
\\ \hline
  \end{tabular}
  \caption{Example simulation in the toothbrushing task. The goal in the shown sub-task is to turn on the tap, take the toothbrush from either the surface or the cup, and wet the brush.}
  \label{TAB:TOOTHRESULTS}
\end{table}

\subsubsection{Real-life Results on Hand Washing}

\begin{figure}
{\tiny
\begin{tabular}{lcr}
 \hline
 a)~~~(2 seconds) & & (frame 24) \\
 \includegraphics[width=.33\textwidth]{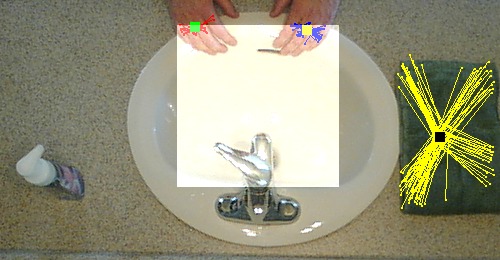} & \includegraphics[width=.35\textwidth]{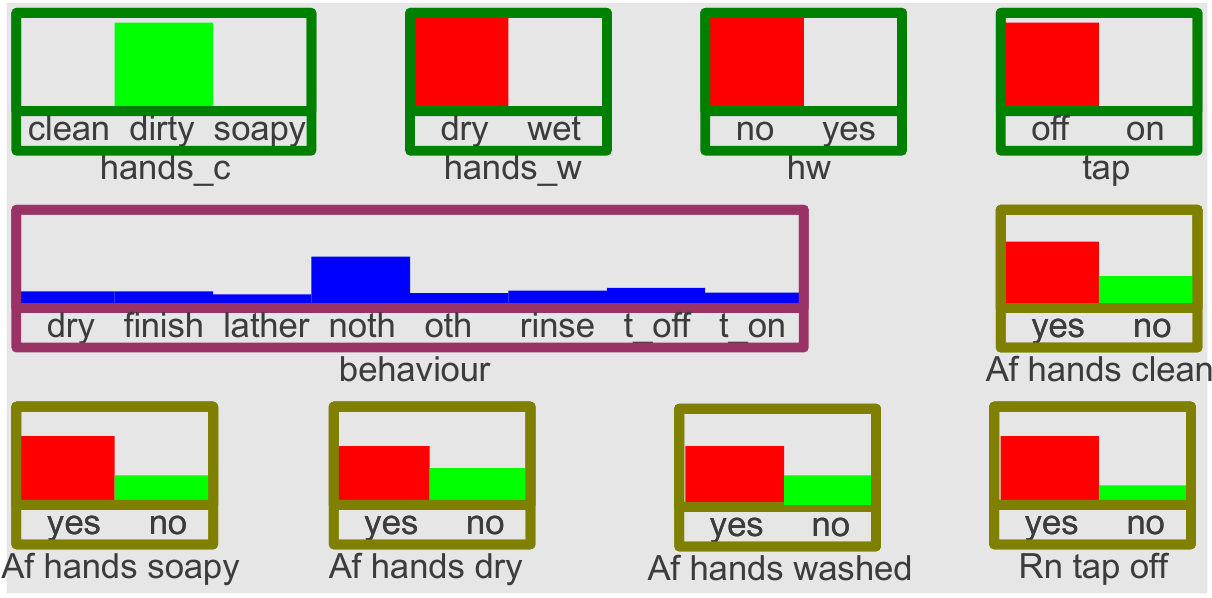} & \includegraphics[width=.24\textwidth]{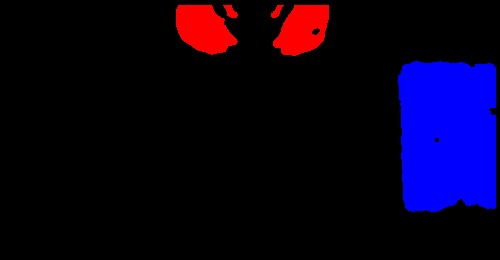}\\ \hline
 b)~~~(6 seconds) & & (frame 82) \\
 \includegraphics[width=.33\textwidth]{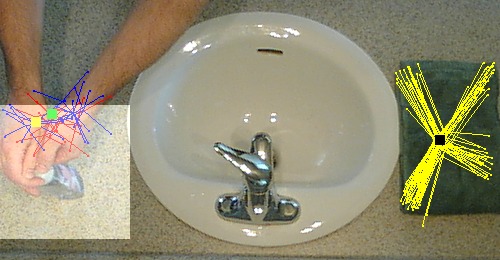} & \includegraphics[width=.35\textwidth]{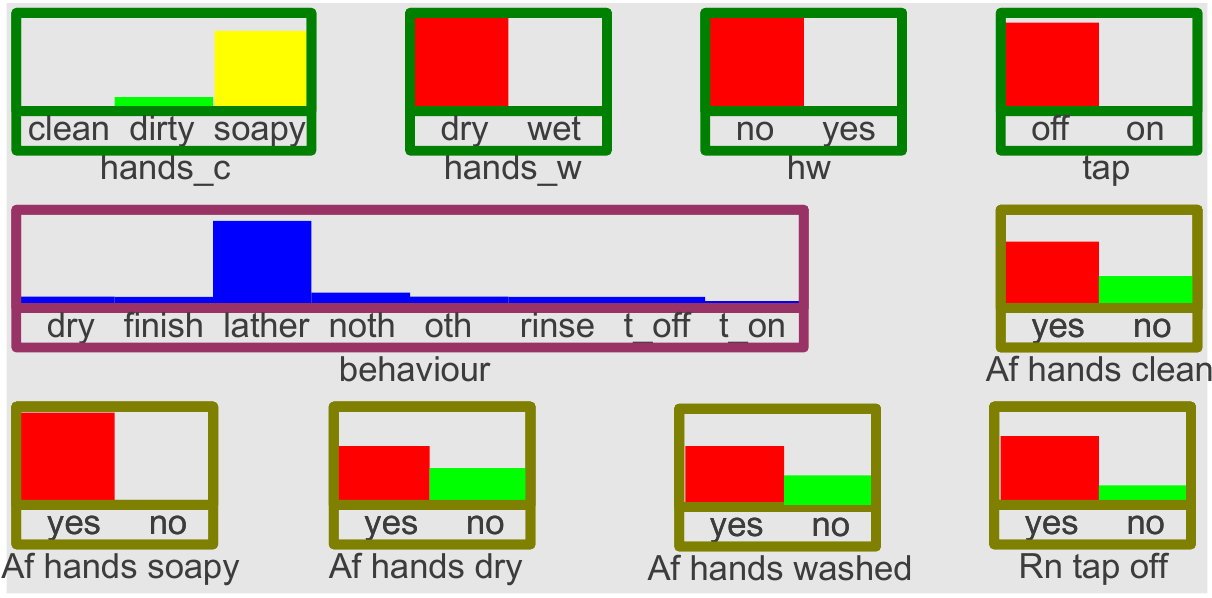} & \includegraphics[width=.24\textwidth]{skin00024.jpg}\\ \hline
 c)~~~(11 seconds) & & (frame 162) \\
 \includegraphics[width=.33\textwidth]{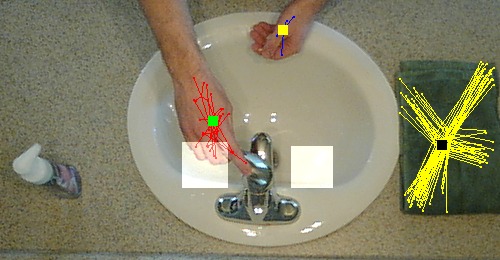} & \includegraphics[width=.35\textwidth]{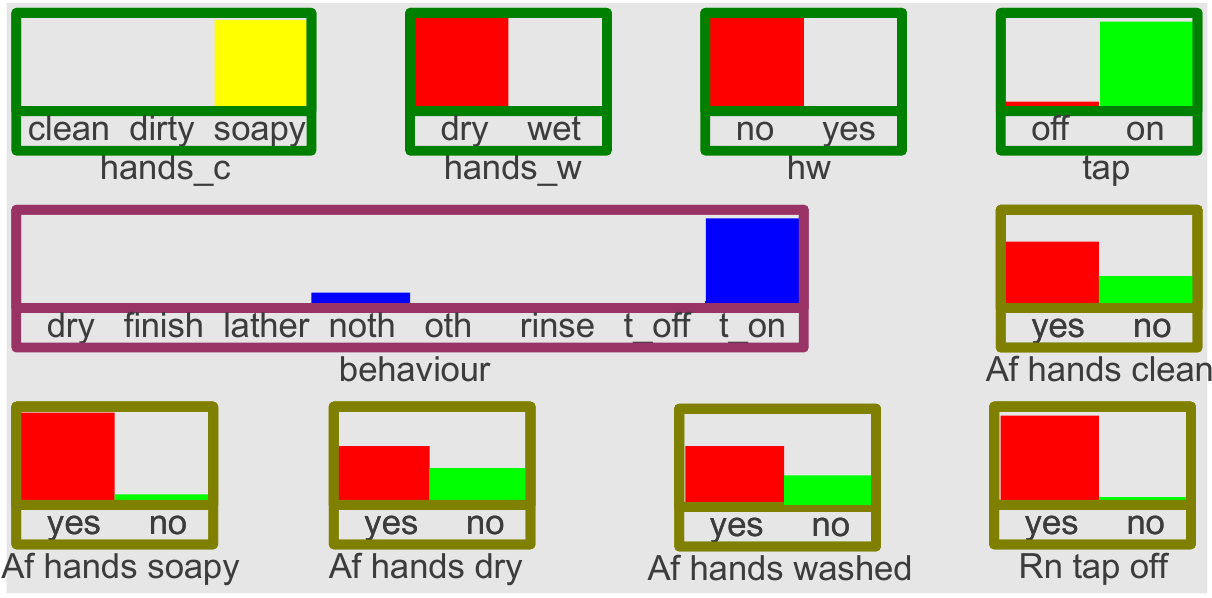} & \includegraphics[width=.24\textwidth]{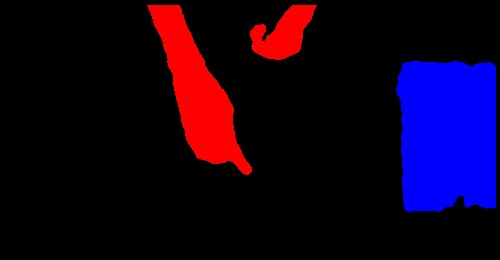}\\ \hline
 d)~~~(18 seconds) & & (frame 260) \\
 \includegraphics[width=.33\textwidth]{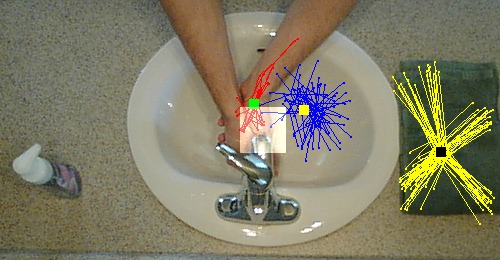} & \includegraphics[width=.35\textwidth]{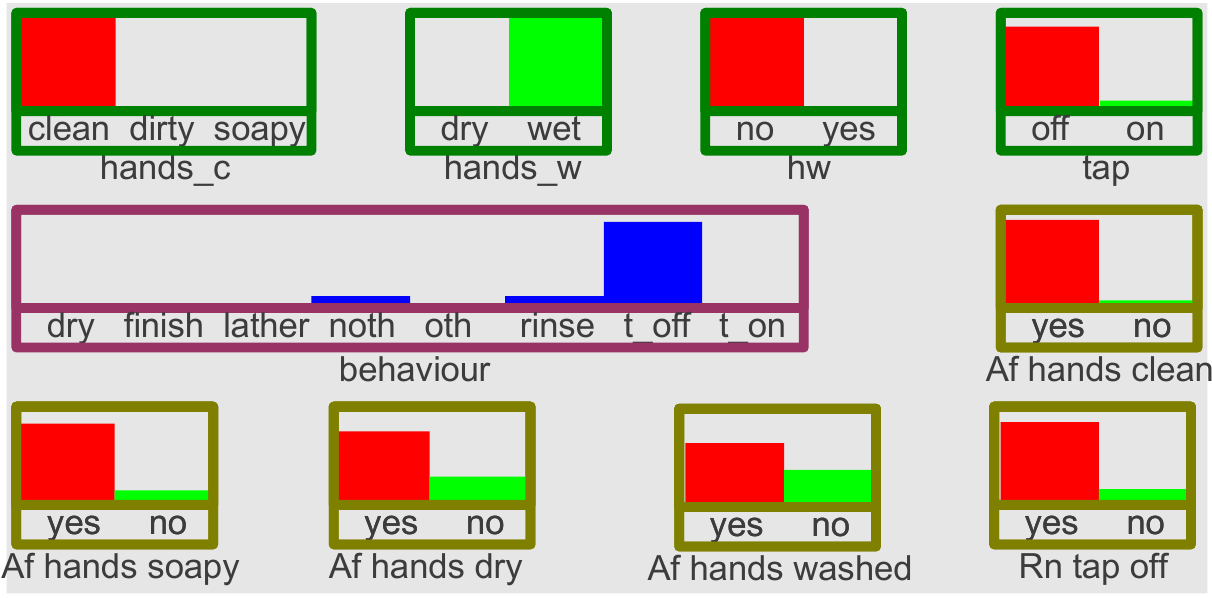} & \includegraphics[width=.24\textwidth]{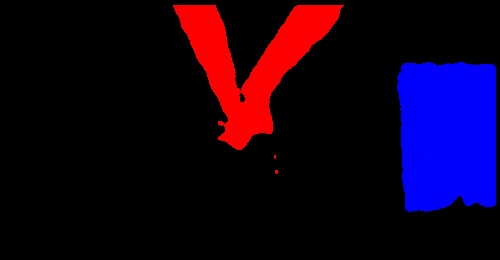}\\ \hline
 e)~~~(25 seconds) & & (frame 354) \\
 \includegraphics[width=.33\textwidth]{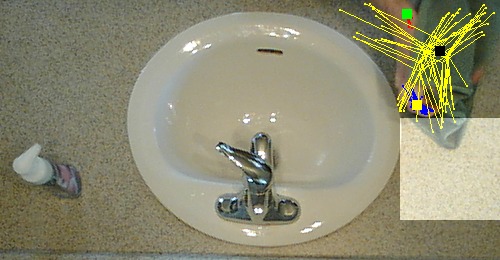} & \includegraphics[width=.35\textwidth]{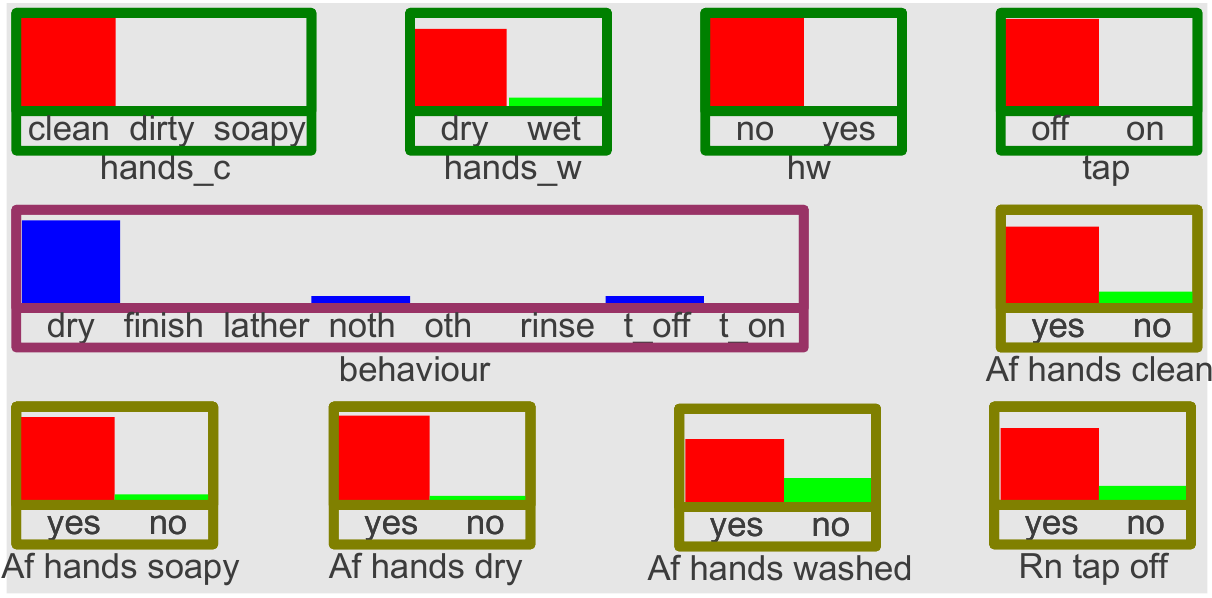} & \includegraphics[width=.24\textwidth]{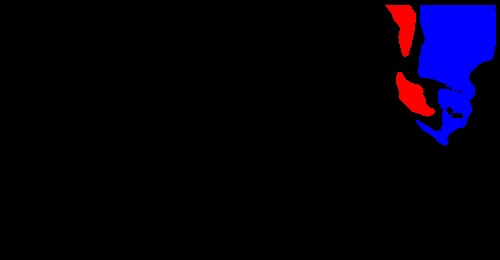}\\ \hline
 f)~~~(32 seconds) & & (frame 465) \\
 \includegraphics[width=.33\textwidth]{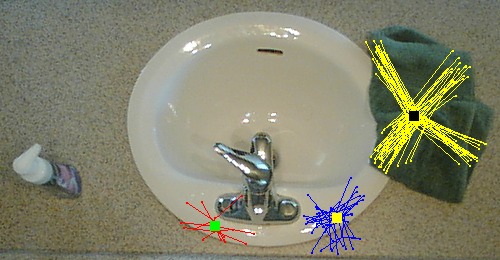} & \includegraphics[width=.35\textwidth]{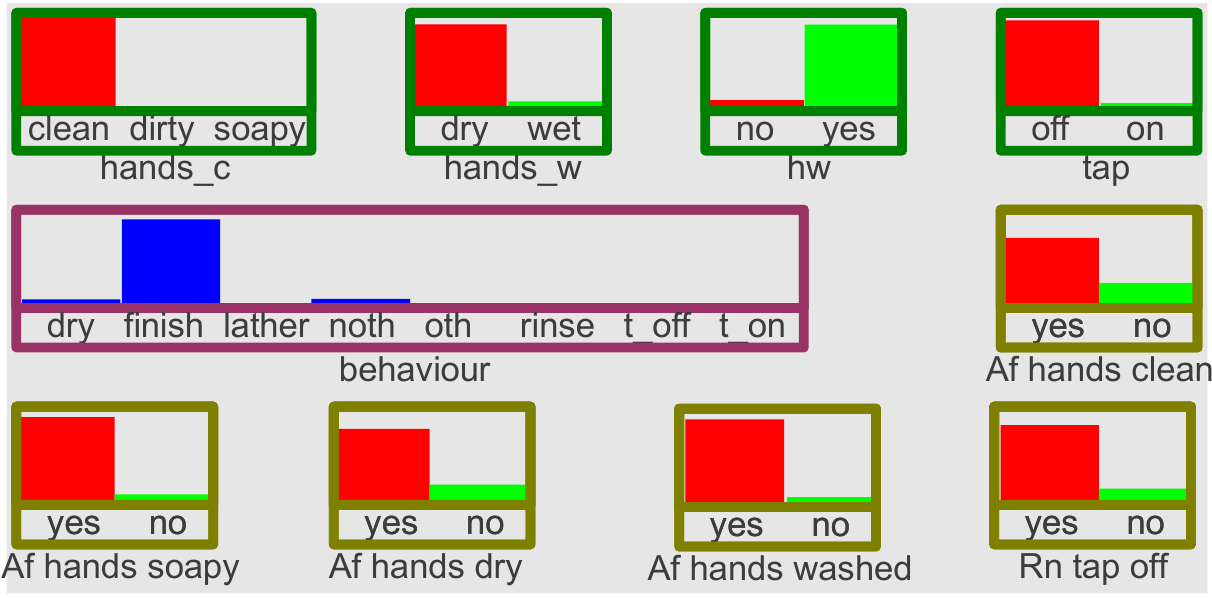} & \includegraphics[width=.24\textwidth]{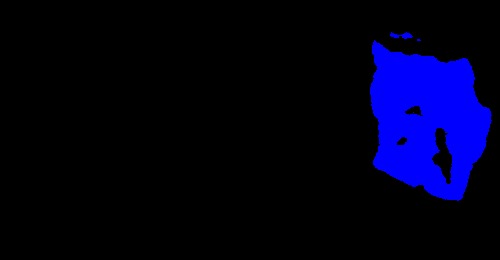}\\ \hline
\end{tabular}}\caption{Key frames from the real-life hand washing experiment using the exact POMDP model that was designed by SC in the case study in Section~\ref{SEC:SYSUSE} and simulated in Table~\ref{TAB:HANDRESULTS}, showing (left) the overhead video and computer vision hand/towel trackers (centre) marginal probabilities over task features, client abilities and behaviours (right) color segmentation image.}
\label{FIG_HWACTOR}
\end{figure}

The same POMDP and policy generated for the handwashing task simulations reported in Section~\ref{S_HW_SIMUL} were used in real-world trials using the COACH system \citep{Boger05b,Hoey10b}. The trials were run with SC impersonating an older adult with mild-to-moderate dementia. Figure~\ref{FIG_HWACTOR} shows the client's progression through the task at key frames in the trial. Each row of this figure shows the relevant belief state(s) in the center, and two images on either side showing how the computer vision system is tracking the hands (see \citep{Hoey10b} for details on these aspects). On the left, we see the overhead video used by the COACH system, overlaid with the (particle-filter based) trackers for hands and towel, and with the regions the hands are in highlighted. On the right, we see the color segmented image that is used by the tracker.  For the trial, the policy is initialized when the client approached the sink (Figure~\ref{FIG_HWACTOR}-a), and the system prompts the client to get soap. The system correctly identifies that the client gets soap (Figure~\ref{FIG_HWACTOR}-b), and prompts to turn on the tap. The client turns on the tap (Figure~\ref{FIG_HWACTOR}-c) and the system prompts the user to rinse the soap off his hands. After scrubbing his hands for a short period of time, the system follows along as he rinses off the soap and turns off the tap (Figure~\ref{FIG_HWACTOR}-d) and the user is prompted to dry his hands. When he dries his hands (Figure~\ref{FIG_HWACTOR}-e) notifies the client that he has completed the task. After leaving the wash basin (Figure~\ref{FIG_HWACTOR}-f), the user is again notified that the task has been completed. This second notification is unnecessary, and can be fixed by reassigning a slightly higher cost for the final prompt. 

\subsection{Factory Assembly Task}

In this example, workers with a variety of intellectual and developmental disabilities are required to complete an assembly task of a `Chocolate First Aid Kit'.  This task is completed at a workstation 
that consists of five input slots, an assembly area, and a completed area.  The input slot contain all of the items necessary to perform kit assembly-specifically the main first aid kit container (white bin), and four different candy containers that need to be placed into specific locations within the kit container.  The IU analysis was completed based on five videos of a specific adult worker (who has Down's Syndrome) completing this assembly task with a human job coach present. Each video was 2-3 minutes in length. The worker was assessed with a moderate-to-mild cognitive impairment and was able to follow simple instructions from a job coach.  The IU analysis broke this task into 6 required steps: 1) prepare white bin; 2) place in bin chocolate bottle 1; 3) place in bin chocolate box 1; 4) place in bin chocolate box 2; 5) place in bin chocolate bottle 2; and 6) finish output bin and place in completed area.  Steps 2, 3, 4, and 5 can be completed in any order.  Through a hierarchical task analysis \citep{Stammers91} each of these steps were further broken down into sub-steps.  
Table~\ref{TAB:FACTORYASSEMBLYIU} is an example IU analysis for step 2.
\begin{table}
  \centering
  {\small
  \begin{tabular}{|l|l|l|l|l|}
\hline
IU & Goals & Task States & Abilities & Behaviours \\ \hline
1  & Final 				& bottle1\_in\_other & Rn: slot\_orange\_empty & Fill slot orange \\
   & assemble chocolate bottle 1 	& slot\_orange\_empty & & \\ \hline
2  & Final 				& bottle1\_in\_slot & Rn: bottle1\_in\_slot & Move bottle1 from slot \\
   & assemble chocolate bottle 1 	& slot\_orange\_full & & \\ \hline
3  & Final 				& bottle1\_in\_hand & Rl: bottle1\_in\_hand & Move bottle 1 \\
   & assemble chocolate bottle 1 	& slot\_orange\_full & & to whitebin \\ \hline
4  & Final 				& bottle1\_in\_whitebin & Af: bottle1\_correctposition & Alter bottle 1 \\
   & assemble chocolate bottle 1 	& slot\_orange\_full & & to correct position \\ \hline
  \end{tabular}
  }
  \caption{IU analysis of step 2 in the factory assembly task. Step 2 of this task requires the client to fill in the orange slot with bottles when it is empty (the orange slot is the place from which bottles are taken and moved to the white bin). When the orange slot contains bottles, the client has to take one bottle, move it to the white bin, and then make sure that the bottle is in the correct position in the white bin.}
  \label{TAB:FACTORYASSEMBLYIU}
\end{table} 

Policies 
were generated for each of the required assembly steps and were simulated by the designer for three different types of clients: mild, moderate, and severe cognitive impairment.   Figure~\ref{TAB:FACTORYASSEMBLYRESULTS} is the output of sample timestamps for step 2 for a client with severe cognitive impairment. Again,
\commentout{
\begin{wrapfigure}{l}{3cm}
  \centering
  \includegraphics[scale=0.15]{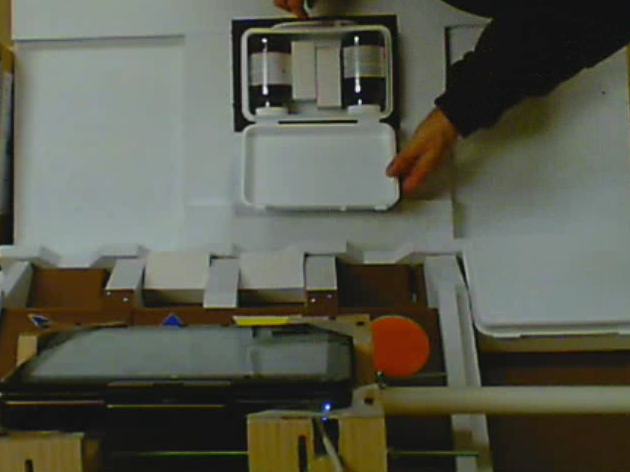}
  \caption{Workstation in the factory assembly task.}
  \label{FIG:FACTORYASSEMBLY}
\end{wrapfigure}
} probabilities of the belief state are represented as the height of bars in corresponding columns of each time step. In this specific example, the system is more active in its prompting based on the fact that the client is assumed to have diminished abilities with respect to the different aspects that needs to be completed.  For example (t=1), the worker has deteriorating ability to recognize that the slot that holds the required chocolate bottle is empty.  As such, the system correctly prompts the worker to recognize that the slot is empty and needs to be filled.  In another example (t=5), the system recognizes that the worker has not placed the bottle in its correct location in the white bin, and provides a prompt for the person to recall that the bottle needs to be in that position in order to reach the final goal state.  When the worker does not respond to this prompt, the system decides (t=6) to play a different, more detailed, prompt
(a prompt related to the affordance ability).
\begin{figure}[tbh]
  \setlength{\tabcolsep}{1pt}
  \centering
  \begin{tabular}{|r|cc|cccccc|cccccc|ccccc|c|}\hline
  & \multicolumn{2}{c|}{Observations} & \multicolumn{6}{c|}{Task} & \multicolumn{6}{c|}{Behaviour} & \multicolumn{5}{c|}{Ability} & \\ \cline{2-20}
  \begin{sideways}Time step, t\end{sideways}
  &\begin{sideways}slot\_orange\_sensor\end{sideways}
  &\begin{sideways}bottle1\_position\_sensor\end{sideways}
  &\begin{sideways}bottle1\_position\_in\_hand\end{sideways}
  &\begin{sideways}bottle1\_position\_in\_whitebin\end{sideways}
  &\begin{sideways}bottle1\_position\_in\_whitebin\_pos1\end{sideways}
  &\begin{sideways}bottel1\_position\_other\end{sideways}
  &\begin{sideways}bottle1\_position\_in\_slot\_orange\end{sideways}
  &\begin{sideways}slot\_orange\_empty\end{sideways}
  &\begin{sideways}other\end{sideways}
  &\begin{sideways}nothing\end{sideways}
  &\begin{sideways}move\_bottle1\_to\_whitebin\end{sideways}
  &\begin{sideways}move\_bottle1\_from\_slot\end{sideways}
  &\begin{sideways}alter\_bottle1\_to\_pos1\end{sideways}
  &\begin{sideways}fill\_slot\_orange\end{sideways}
  &\begin{sideways}Rl\_bottle1\_position\_in\_whitebin\_pos1\end{sideways}
  &\begin{sideways}Rn\_slot\_orange\_empty\end{sideways}
  &\begin{sideways}Af\_alter\_bottle1\_to\_pos1\end{sideways}
  &\begin{sideways}Rl\_bottle1\_position\_in\_hand\end{sideways}
  &\begin{sideways}Rn\_bottle1\_position\_in\_slot\_orange\end{sideways}
  & \begin{sideways}\makecell{System Action\\(prompt for ability specified)}\end{sideways}
\\ \hline
  0 &- &-
  &\rule{3mm}{0.01mm} &\rule{3mm}{0.1mm} &\rule{3mm}{0.01mm} &\rule{3mm}{.4mm} &\rule{3mm}{1.6mm} &\rule{3mm}{.4mm}
  &- &- &- &- &- &-
  &\rule{3mm}{1.4mm} &\rule{3mm}{1.4mm} &\rule{3mm}{1.4mm} &\rule{3mm}{1.4mm} &\rule{3mm}{1.4mm}
  & Rn bottle1 in slot\\
  1 &empty &other
  &\rule{3mm}{0.01mm} &\rule{3mm}{0.01mm} &\rule{3mm}{0.01mm} &\rule{3mm}{1.46mm} &\rule{3mm}{.52mm} &\rule{3mm}{1.9mm}
  &\rule{3mm}{0.32mm} &\rule{3mm}{0.32mm} &\rule{3mm}{0.32mm} &\rule{3mm}{0.34mm} &\rule{3mm}{0.32mm} &\rule{3mm}{0.08mm}
  &\rule{3mm}{0.48mm} &\rule{3mm}{0.18mm} &\rule{3mm}{0.82mm} &\rule{3mm}{0.48mm} &\rule{3mm}{1.6mm}
  & Rn slot empty \\
  2 &empty &other
  &\rule{3mm}{0.01mm} &\rule{3mm}{0.01mm} &\rule{3mm}{0.01mm} &\rule{3mm}{1.8mm} &\rule{3mm}{0.18mm} &\rule{3mm}{1.98mm}
  &\rule{3mm}{0.32mm} &\rule{3mm}{0.32mm} &\rule{3mm}{0.32mm} &\rule{3mm}{0.32mm} &\rule{3mm}{0.32mm} &\rule{3mm}{0.02mm}
  &\rule{3mm}{0.30mm} &\rule{3mm}{0.54mm} &\rule{3mm}{0.64mm} &\rule{3mm}{0.30mm} &\rule{3mm}{0.52mm}
  & Rn slot empty \\
  3 &full &in\_slot\_orange
  &\rule{3mm}{0.01mm} &\rule{3mm}{0.01mm} &\rule{3mm}{0.01mm} &\rule{3mm}{0.01mm} &\rule{3mm}{1.98mm} &\rule{3mm}{0.01mm}
  &\rule{3mm}{0.01mm} &\rule{3mm}{0.01mm} &\rule{3mm}{0.01mm} &\rule{3mm}{0.01mm} &\rule{3mm}{0.01mm} &\rule{3mm}{1.98mm}
  &\rule{3mm}{0.26mm} &\rule{3mm}{1.92mm} &\rule{3mm}{0.6mm} &\rule{3mm}{0.26mm} &\rule{3mm}{0.3mm}
  & Rn bottle1 in slot \\
  4 &full &in\_hand
  &\rule{3mm}{1.96mm} &\rule{3mm}{0.01mm} &\rule{3mm}{0.01mm} &\rule{3mm}{0.01mm} &\rule{3mm}{0.4mm} &\rule{3mm}{0.01mm}
  &\rule{3mm}{0.01mm} &\rule{3mm}{0.01mm} &\rule{3mm}{0.01mm} &\rule{3mm}{1.96mm} &\rule{3mm}{0.01mm} &\rule{3mm}{0.01mm}
  &\rule{3mm}{0.26mm} &\rule{3mm}{0.58mm} &\rule{3mm}{0.58mm} &\rule{3mm}{0.26mm} &\rule{3mm}{1.9mm}
  & Rl bottle1 in hand \\
  5 &full &in\_whitebin
  &\rule{3mm}{0.4mm} &\rule{3mm}{1.96mm} &\rule{3mm}{0.01mm} &\rule{3mm}{0.01mm} &\rule{3mm}{0.01mm} &\rule{3mm}{0.01mm}
  &\rule{3mm}{0.01mm} &\rule{3mm}{0.01mm} &\rule{3mm}{1.96mm} &\rule{3mm}{0.01mm} &\rule{3mm}{0.01mm} &\rule{3mm}{0.01mm}
  &\rule{3mm}{0.26mm} &\rule{3mm}{0.32mm} &\rule{3mm}{0.58mm} &\rule{3mm}{1.9mm} &\rule{3mm}{0.58mm}
  & Af bottle1 to pos1\\
  6 &full &in\_whitebin
  &\rule{3mm}{0.01mm} &\rule{3mm}{1.96mm} &\rule{3mm}{0.4mm} &\rule{3mm}{0.01mm} &\rule{3mm}{0.01mm} &\rule{3mm}{0.01mm}
  &\rule{3mm}{0.32mm} &\rule{3mm}{0.32mm} &\rule{3mm}{0.32mm} &\rule{3mm}{0.32mm} &\rule{3mm}{0.4mm} &\rule{3mm}{0.32mm}
  &\rule{3mm}{0.08mm} &\rule{3mm}{0.26mm} &\rule{3mm}{1.46mm} &\rule{3mm}{0.58mm} &\rule{3mm}{0.32mm}
  & Rl botte1 in bin \\
  7 &full &in\_whitebin
  &\rule{3mm}{0.01mm} &\rule{3mm}{1.9mm} &\rule{3mm}{0.1mm} &\rule{3mm}{0.01mm} &\rule{3mm}{0.01mm} &\rule{3mm}{0.01mm}
  &\rule{3mm}{0.32mm} &\rule{3mm}{0.32mm} &\rule{3mm}{0.32mm} &\rule{3mm}{0.32mm} &\rule{3mm}{0.08mm} &\rule{3mm}{0.32mm}
  &\rule{3mm}{1.42mm} &\rule{3mm}{0.26mm} &\rule{3mm}{0.38mm} &\rule{3mm}{0.32mm} &\rule{3mm}{0.26mm}
  & Af bottle1 to pos1 \\
  8 &full &in\_whitebin\_pos1
  &\rule{3mm}{0.01mm} &\rule{3mm}{0.22mm} &\rule{3mm}{1.88mm} &\rule{3mm}{0.01mm} &\rule{3mm}{0.01mm} &\rule{3mm}{0.01mm}
  &\rule{3mm}{0.4mm} &\rule{3mm}{0.26mm} &\rule{3mm}{0.4mm} &\rule{3mm}{0.4mm} &\rule{3mm}{1.44mm} &\rule{3mm}{0.4mm}
  &\rule{3mm}{1.06mm} &\rule{3mm}{0.26mm} &\rule{3mm}{1.74mm} &\rule{3mm}{0.26mm} &\rule{3mm}{0.26mm}
  & do nothing \\
  \hline
  \end{tabular}
  \caption{Example simulation in the factory assembly task. The goal in the shown sub-task is to take the bottle, named bottle 1, from the orange slot and to place the bottle in the white bin in pos1.}
  \label{TAB:FACTORYASSEMBLYRESULTS}
\end{figure}

\section{Related Work}

Below, we relate our work to the existing research on knowledge engineering for planning which is the main target of this paper and also to statistical relational learning which motivates our relational solution.

\subsection{Knowledge Engineering for Planning}


We start this section with a reference to a more general approach of reinforcement learning (RL) which represents a broader class of planning problems where available domain knowledge is sufficient for only a partial formulation of the problem and the planning algorithm has to estimate the missing elements using simulation \citep{bertsekas96neuro-dynamic}. A recent survey of the existing tools and software for engineering RL problem specifications presented in \citep{kovacs11RLsoft} shows that engineering of RL domains is in most cases done either by re-implementation of required algorithms and domains/simulators or by partial re-use of the existing source code in the form of libraries or repositories, which means that engineering of RL domains is a direct implementation problem. There are no existing out of the box, domain independent environments where the specification of the RL problem would be reduced to the specification of the domain in a specific domain definition language.

Existing tools are far more advanced with this regard in the area of planning (both symbolic and MDP-based decision theoretic planning) where planners have been made publicly available and these planners can read the planning problem in a specific language (e.g., a variation of STRIPS is used in Graphplan \citep{blum97fast} or SPUDD in the SPUDD planner \citep{Hoey99}) directly and no coding of the planner is required from the user. Such planners are called domain independent planners \citep{ghallab04automated}, and it is sufficient for the designer of the planning domain to know the planner's domain definition language and specify the domain in that language.

The process of defining the planning domain in the language of the planner can be still tedious or challenging for less experienced users and there has been past research which aims at providing tools which can help in the process of defining the planning problems \citep{edelkamp05modplan,simpson07GIPO,vacquero05simple}. The difference between these tools and our work is that they assume that the planning problem has been already identified by the designer and the goal of the tool is to help the designer in formalising the description using the specific formal language which is associated with the planner. Our work introduces a further separation between the designer and the planning domain specification by reducing the interaction between the designer and our tool to \textit{the translation process} where the designer performs a psychological IU analysis of the task, enters the result into the database using a standard web-based interface, and then the planning problem definition of the prompting system is generated automatically from the content of the database. Actually, the designer does not have to be even fully aware that what he is doing in the database is the specification of the AI planning domain which corresponds to the amechanistic property of the system.

\subsection{Probabilistic Relational Planning}

First-order models allow for a compact representation of planning domains since the early days of symbolic planning where the STRIPS formalism is a good early example \citep{fikes71strips}. Various alternative and extended formulations were proposed such as PDDL \citep{helmert09pddl} and its probabilistic version PPDDL \citep{younes04ppddl}. Even though they are relational in their nature, it is not always easy to specify certain dependencies in the modelled domain. This was probably the reason why representations based on Dynamic Bayesian Networks (DBNs) were introduced \citep{Dean89}. Classical examples are SPUDD and SymbolicPerseus languages \citep{Hoey99,PoupartThesis04}. They lack however first-order principles and things which are standard in STRIPS, such as parametrised actions, were not possible in existing DBN-based representations, until recently RDDL combined ideas from models based on DBNs and variations of STRIPS \citep{sanner11RDDL}. This makes RDDL an extended version of SPUDD/SymbolicPerseus \citep{Hoey99,PoupartThesis04} with parametrised actions plus several other extensions. It is interesting to note that RDDL applied the model of actions being DBN variables, which allows for concurrency but which exists also in prompting systems modelled using SPUDD, where client's actions - behaviours in our model - are variables in the DBN (see Figure~\ref{FIG:DB} for details).

We now look how the above formalisms are related to our work. RDDL becomes very convenient because conditional probability distributions and actions can be modelled in a compact way due to their parametrisation (like in STRIPS). Even though SPUDD/SymbolicPerseus language which we used does not have this property, we moved this task to the database and domain generator software which grounds actions on the fly and saves them to the SPUDD/SymbolicPerseus representation. The RDDL based planner would need to ground actions at the planning level, and we do this at the intermediate representation level.

Initially, relational formalisms were used for a compact representation of the planning problem only, and planners which were used for actual planning were grounded and did not exploit relational structure in the domain. In the last decade, some solid - at least theoretically - work has emerged which aims at using relational representations explicitly during planing in the so called lifted planning or first-order planning \citep{boutilier01symbolic,kersting04bellman,sanner08thesis}. The work of this paper does not aim at improving any specific POMDP planner, either lifted or grounded. Instead, the use of relational modelling in our paper helps formulate the planning domain, and the fact that our methodology is based on a translation of a psychological model in conjunction with relational modelling makes our tool accessible for designers who are not experts in POMDP planning.

\subsection{Statistical Relational Learning}
The idea of Statistical Relational Learning (SRL) is to extend propositional probabilistic models (e.g., Bayesian or decision networks) with the concept of objects, their attributes, and relations between objects.  Two types of representations are common in the SRL research: rule-based \citep{sato95PRISM} and frame-based \citep{GetoorSRL07}. We base our work on the example of the second type which is a Probabilistic Relational Model \citep{getoor07prm}, because this model has an inherent connection with a tabular representation of data which is found in relational databases. 

The discussion of our model as a PRM established a natural connection with relational databases. It defines objects, relations between objects and templates for probability distributions which can use aggregating operators in order to deal with varying numbers of parameters. In our case, the parametrisation of objects and possible relations between objects was also required and we achieved it by defining attributes of main domain objects, also as objects. The resulting model encodes a template for specifying relational models, which when grounded, represent a specific type of POMDPs  model assistance tasks.

Similar requirements arise in statistical relational learning where analogous templates are necessary in order to deal with `deep knowledge transfer'. This is the case in \citep{davis08deeptransfer} where general transferable knowledge is extracted. The second-order logic model defines the hypothesis space for machine learning algorithms in the same way as in our frame-based model the template for assistance POMDPs is defined. If the logic- or rule-based SRL system does not have to cope with `deep knowledge transfer', first-order models are usually sufficient because they concentrate on repeated structures which is satisfactory in order to provide the use of objects and relations between those objects \citep{laskey08mebn,getoor07prm,koller97OOBNs}. It is worth remembering that standard plate models also allow modelling repeated structures.

Lifted planning is still a challenge and currently most planners are grounded planners. A similar situation exists in statistical relational learning where most actual reasoning or learning happens after grounding the model \citep{domingos09MLNbook} though recent work shows promising advancements in lifted reasoning \citep{broeck11liftedInference,gogate11prob_theorem_proving}. Regardless of specific architectures for planning or statistical relational learning, i.e., grounded or lifted, relational models are very powerful in designing probabilistic models.

\section{Future Work}\label{SEC:FUTUREWORK}
The SNAP methodology~\citep{SnapPMC11} was designed to enable non-technical users to specify POMDP prompting systems.  There is an inherent trade-off involved whereby the complexity of the model must be sacrificed for ease of specification.  The database we have presented in this paper implements only the most basic functionality, allowing a designer to specify a prompting system for only the core set of assistive technologies (e.g. the handwashing system or other simple tasks with atomic client behaviours, no exogenous events, and a restricted set of variable influences as shown in Figure~\ref{fig:pomdp}).  In this section we discuss various extensions to this basic model that can be enabled by enlarging the database and making specification of the model more complex.

\subsection{Exogenous events} In the basic model, only the client can change the state of the world via his behaviours. 
However, in some cases the state may be changed by exogenous events, such as the weather, clocks or other persons (e.g. caregivers). For example, we are working on a system to help a person with dementia when they get lost while outside (wandering)~\citep{HoeyFavela12b,HoeyFavela12a}. In this case, the task state may involve external elements such as weather, or proximity of a caregiver. The POMDP models for this domain were obtained using the database system presented in this paper, and additional dependencies were added easily, though the designer must specify some additional probabilities (e.g. the probability of arrival of a caregiver, or the probability the weather will change).

\subsection{Sensing the cognitive state} To apply our model to a dialogue system \citep{frampton09RL4dialogue,williams11dialog}, we need to allow for sensing the cognitive state of the client (e.g.~by asking a question).  One of our projects is to add a dialogue component to an existing prompting system, and we found that system actions were required to ask the client a question about her specific cognitive features. The potential advantage of using our system for generating dialogue POMDPs should be explored more deeply in the future, because it could allow for a straightforward integration of standard slot/filling dialogue systems with those that require the model of the task. Such approaches are still challenging in research on dialogue systems~\citep{frampton09RL4dialogue} and our methodology is addressing this missing link.

\subsection{Sensing the task state} The situation mentioned in the previous point may also extend to the case when the system cannot sense a specific task feature and would need to ask the client a question about the state of that feature. Appropriate system actions would be required that would serve as sensors.


\subsection{Generalised Model of Assistance}
Our methodology and the existing software do not have to be limited to particular types of assistive systems that we presented so far in the paper. In order to show this fact, we briefly discuss here a generalised model of human assistance that could be used in a broad range of applications
 where the cognitive model of the human being is required or useful. 
The first extension to the original model is that a more general cognitive state of the client is required that has a broader meaning than the {\em abilities} used in this paper. 
Another extension is that, in some applications such as tutoring systems~\citep{Murray00}, it is useful to model the cognitive behaviours of the client, and for this reason, the behaviour should be applicable for both physical (i.e., belonging to the environment or real world) and cognitive behaviours of the client.
For example, in a tutoring system, the cognitive behaviour of the client reaching the understanding of a new concept would trigger a change in the cognitive state of the client. The dependency of the cognitive state on the behaviour does not require changes to our system, because the cognitive state that does require this feature could be placed in the same table as the standard task state.


\section{Conclusions}

POMDP models have proven to be powerful for modelling intelligent human assistance \citep{Hoey10b}. Unfortunately, POMDPs, being propositional models, usually require a very labour intensive, manual set-up procedure. Recent work has shown how this can be simplified by using a methodology from the psychology literature called interaction unit (IU) analysis~\citep{SnapPMC11}.  In this paper, we show how the introduction of a probabilistic relational model and a database encoding can greatly simplify and standardise this process. 
We derived a methodology for specifying POMDPs for intelligent human assistance which is motivated by relational modelling in frame-based SRL methods \citep{GetoorSRL07}. The core of our approach is the relational database which the designer populates in order to prepare the deployment of the system for a specific task.

Traditionally, the specification of POMDPs was performed by trained POMDP experts who would encode the problem in a specific notation that the planner can understand. The existing research investigated the problem of how this specification can be done by non-POMDP experts, where the specialised tool helps designing the POMDP. The key feature of such systems is that the designer still explicitly defines the planning problem. In this paper, we implement another way of defining POMDPs through a translation of a psychological model. The designer is doing the psychological task analysis of the task, whereas the system takes the data provided by the designer and automatically translates it into the valid POMDP in the background. This results in a new paradigm that frees the designer from the burden of knowing the POMDP technology. Our design is facilitated by the correspondence between the psychological Interaction Unit (IU) analysis and POMDPs for assistive systems \citep{SnapPMC11}, and by statistical relational modelling in artificial intelligence.

In our system, content provided by the designer of a particular implementation encodes the goals, action preconditions, environment states, cognitive model, client and system actions, as well as relevant sensor models, and automatically generates a POMDP model of the assistance task being modelled. The strength of the database is that it also allows constraints to be specified, such that we can verify the POMDP model is valid for the task. To the best of our knowledge, this is the first time the POMDP planning problem is formalised using a relational database.

We demonstrate the method on three assistance tasks: handwashing and toothbrushing for elderly persons with dementia, and on a factory assembly task for persons with a cognitive disability. This demonstration shows that the system, once designed using the relational approach, can be instantiated to create a POMDP controller for an intelligent human assistance task. The use of the relational database makes the process of specifying POMDP planning tasks straightforward and accessible to non-expert computer users. We have applied this method to other tasks as well, including location assistance and dialogue management for scheduling applications. 

A huge benefit of our methodology for specifying POMDPs is a massive reduction of time required for designing formal POMDP models. A manual coding of the system for handwashing (a smaller task) took over 6 months of work resulting in the system described in~\citep{Boger05b}. Our system allows obtaining such models in no more than several hours of work of a non-POMDP expert.

The problem of formal specification of POMDP models exists in other applications and areas of applied artificial intelligence research. An example where our system could be applied are POMDP-based dialogue systems which often apply POMDPs as a decision making engine, whereas they are usually specified by experts in natural language processes who are not necessarily POMDP experts. We believe that this work shows how the specification of POMDPs can be made accessible for non-experts in various areas of interest.

We currently release our POMDP representation and solution software as an open-source distribution. Upon publication of this paper, we will further release our POMDP generation software under the same distribution. Further, our system is available online\footnote{Web address: \url{http://www.cs.uwaterloo.ca/~mgrzes/snap} with instant access and no email verification required. Readers can also view a short video demonstrating a complete walk-through at \url{http://youtu.be/KHx9zGljkLY}} for testing by interested users using a secure web-based access system. This allows a user to create an account on our system, to provide the data for his/her task, and then to generate a valid POMDP model for his/her task without any software downloads.




\section{Acknowledgements} This research was sponsored by American Alzheimer's Association, the Toronto Rehabilitation Institute (TRI), and by NIDRR as part of the RERC-ACT at ATP Partners, University of Colorado-Denver. The first author was supported by a fellowship from the Ontario Ministry of Research and Innovation.

\bibliographystyle{elsarticle-harv}
\bibliography{refs}  

\end{document}